%% file: acl2025.tex
\pdfoutput=1

\documentclass[11pt,a4paper]{article}
\usepackage[hyperref,final]{acl}
\usepackage{times}
\usepackage{inconsolata}
\usepackage{latexsym}
\usepackage{danudefs}
\usepackage{algorithmic}
\usepackage{algorithm}
\usepackage{color}
\usepackage{booktabs}
\usepackage{xspace}
\usepackage{url}
\usepackage{braket}
\usepackage[T1]{fontenc}
\usepackage[utf8]{inputenc}
\usepackage{graphicx}
\usepackage{subcaption}
\usepackage{tcolorbox}
\usepackage{microtype}
\usepackage{enumitem}
\usepackage{amsmath}
\usepackage{multirow}
\usepackage[english]{babel}
\usepackage{hyperref}
\addto\captionsenglish{%
}
\addto\extrasenglish{%
}

\usepackage{makecell}

\input{myacronyms}

\title{Evaluating the Evaluation of Diversity in Commonsense Generation}

\author{Tianhui Zhang \And
    Bei Peng \\
  University of Liverpool\\
   {\tt \{tianhui.zhang, danushka, bei.peng\}@liverpool.ac.uk} \\ \And
  Danushka Bollegala}

\date{}

\begin{document}
\maketitle

\begin{abstract}
    In commonsense generation, given a set of input concepts, a model must generate a response that is not only commonsense bearing, but also capturing multiple diverse viewpoints.
    Numerous evaluation metrics based on form- and content-level overlap have been proposed in prior work for evaluating the diversity of a commonsense generation model.
    However, it remains unclear as to which metrics are best suited for evaluating the diversity in commonsense generation.
    To address this gap, we conduct a systematic meta-evaluation of diversity metrics for commonsense generation.
    We find that form-based diversity metrics tend to consistently overestimate the diversity in sentence sets, where even randomly generated sentences are assigned overly high diversity scores.
    We then use an \ac{LLM} to create a novel dataset annotated for the diversity of sentences generated for a commonsense generation task, and use it to conduct a meta-evaluation of the existing diversity evaluation metrics.
    Our experimental results show that content-based diversity evaluation metrics consistently outperform the form-based counterparts, showing high correlations with the \ac{LLM}-based ratings.
    We recommend that future work on commonsense generation should use content-based metrics for evaluating the diversity of their outputs.
\end{abstract}

\section{Introduction}
\label{sec:intro}

Commonsense reasoning---the ability to make plausible assumptions about ordinary scenarios---is a core requirement for robust \ac{NLG} systems~\cite{CommonGen}.
In the task of \ac{GCR}, an \ac{NLG} model is expected to generate sentences that are both \emph{quality-bearing} (i.e. logically coherent and commonsense-aware) and \emph{diverse} (i.e. offering varied perspectives on the same input concepts)~\cite{dimongen,yu:2022:diversifying,hwang:2023:knowledge}.

\begin{figure}[t]
\centering
\includegraphics[width=1.0\linewidth]{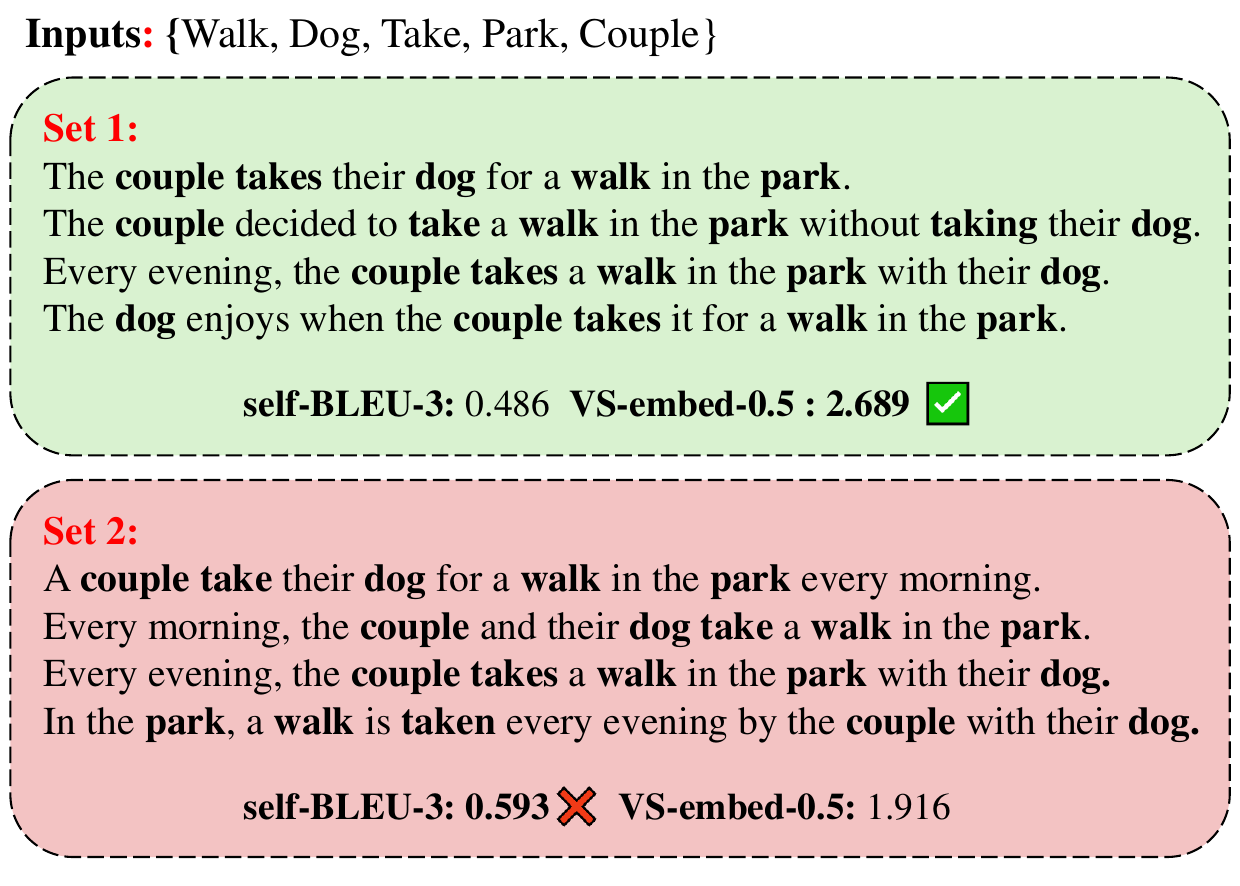}
\caption{An example from the CommonGen~\citep{CommonGen} dataset comparing two sets of generated sentences. \texttt{self-BLEU-3} indicates Set-2 to be more diverse, which simply repeats near-identical paraphrases. In contrast, \texttt{\ac{VS}-embed-0.5}  aligns well with the notion of meaningful textual diversity.}
\label{fig:diversity_example}
%\vspace{-5mm}
\end{figure}

While recent neural architectures have significantly improved the quality of commonsense generation, reliably evaluating the diversity of generated outputs remains an open challenge.
Quality evaluation typically relies on comparing generated outputs against a set of human-written reference sentences using metrics such as BLEU~\cite{bleu}, ROUGE~\cite{rouge}, or SPICE~\cite{spice}.
A \ac{GCR} method that produces outputs that have a high overlap with human-written reference sentences is considered to be of \emph{high quality}.
%\textbf{form} (i.e. surface-level diversity is measured using for example $n$-gram overlap) or \textbf{content} (i.e. meaning-level diversity measured, for example, using the semantic similarity computed using sentence embeddings).
In contrast, diversity is assessed by comparing the outputs among themselves. 
A variety of diversity metrics have been proposed~\cite{li:2016:distinct,zhang:2024:diversity} and can be broadly categorised into two groups: \textbf{form-based} vs. \textbf{content-based}. 
Form-based diversity metrics such as self-BLEU~\cite{zhu:2018:selfbleu} and distinct~\cite{li:2016:distinct}, measure the token/word overlap between pairs of sentences using  $n$-grams,
whereas content-based diversity metrics such as self-CosSim~\citep{cox2021directed} and Vendi-Score~\cite{vendi:2022} capture semantic variations using sentence embeddings.

A central question arises: \emph{Which diversity metrics best capture meaningful variations in commonsense generation, and under what conditions?}
For instance, as shown in \autoref{fig:diversity_example}, given the five input concepts  \emph{walk, dog, take,  park} and  \emph{couple}, a \ac{GCR} method must produce sentences that contain all of the input concepts and their diverse commonsense relations.
Although both Set-1 and Set-2 contain commonsense-making sentences covering all input concepts, Set-2 contains direct paraphrases or random word-order shuffles.
Consequently, Set-2 is less diverse compared to Set-1.
However, the form-based diversity metrics (e.g. self-BLEU3) assign high diversity scores to Set-2 than to Set-1, overestimating the diversity in \ac{GCR}.
As we later see in our meta-evaluations (\autoref{sec:high-vs-low-eval}), form-based diversity metrics tend to assign high diversity scores even for randomly generated nonsensical sentences, which is counter-intuitive.
On the other hand, content-based diversity metrics (e.g. \ac{VS}-embed-0.5) seem less susceptible to such issues and correctly predict Set-1 to have a higher diversity than Set-2.

% We conduct a meta-evaluation of diversity metrics to address the RQ. However, large-scale human evaluation of diversity is difficult (Tevet). Thus, we use LLM for annotation. Explain the protocol.

We conduct a comprehensive meta-evaluation of 12 diversity metrics for \ac{GCR} using three standard \ac{GCR} datasets.
For this purpose, we create a large-scale diversity-annotated dataset.
Prior work studying diversity~\cite{tevet2021evaluating} in \ac{NLG} has shown difficulty in obtaining reliable diversity ratings via crowdsourcing.
However, \citet{zhang:2024:diversity} showed that \acp{LLM} could be used to evaluate the diversity in \ac{GCR} with a moderate-level of agreement with linguistically trained human annotators.
We follow their work and create a dataset where an \ac{LLM} provides a pairwise preference rating for two sets of sentences covering the same input concepts.
A human evaluation on a subset of our dataset shows the \ac{LLM}-based diversity ratings to be well-aligned with the human judgments  with an average accuracy of 80.6\%.

Next, we measure the pairwise preference agreement between the \ac{LLM}-based ratings and diversity metrics for high vs. low quality generations.
%For example, for the two sets of sentences shown in \autoref{fig:diversity_example}, the \ac{LLM} correctly predicts Set-1 to be more diverse compared to Set-2.
%To study the relationship between the reliability of a diversity metric and the quality of the sentences being evaluated, we analyse the diversity ratings for high vs. low quality generations.
We find that, 
\begin{enumerate}
    \item Form-based diversity metrics produce reliable evaluations for high quality generations, but often fail to distinguish genuine diversity for the lower-quality generations, and
    \item Content-based metrics produce consistently reliable evaluations for both high and low quality generations.
\end{enumerate}

Our code and data are available at \url{https://github.com/LivNLP/Evaluating-Diversity-Metrics}.

\section{Related Work}
\label{sec:related}
%Diverse Commonsense Generation, I'm not sure if we need to put it into the related work as this is not our focus, we only evaluate, maybe we just mention they are not well evaluated on diversity?

\paragraph{Diversity in NLG:}
Diverse output generation is a critical requirement for many \ac{NLG} applications ~\citep{tevet2021evaluating} such as storytelling~\citep{li:2018:story}, question generation~\citep{pan2019recent} and machine translation~\citep{shen:2019:mixture}.
Strategies proposed for improving diversity in \ac{NLG} include sampling methods that prune the probability distribution over the next-token predictions such as nucleus sampling~\citep{holtzman:2019:sample} and top-$k$ sampling~\citep{fan:2018:hierarchical}.
Setting high temperature for the decoder~\citep{peeperkorn:2024:temperature} can sometimes increase the diversity in the generated output but must be done with care as it can decrease the quality~\citep{zhang:2024:diversity}.
%\ac{MoE} approach is introduced to diversify translation outputs~\citep{shen:2019:mixture}.
%These generations are typically evaluated using diversity-oriented metrics (e.g., self-BLEU~\citep{zhu:2018:selfbleu}), which were originally developed for other \ac{NLG} tasks. It remains to be seen whether these metrics are equally effective in the context of \ac{GCR}, where commonsense-bearing is the core.

\paragraph{Diversity in \ac{GCR}:}
Diversification in \ac{GCR} presents an additional layer of complexity because we must generate both diverse as well as commonsense bearing outputs.
Datasets such as CommonGen~\citep{CommonGen} and DimonGen~\citep{dimongen} provide a set of concepts and a set of sentences that describe various commonsense relations among those concepts, while ComVE~\citep{semeval} requires a \ac{GCR} method to explain why a given counterfactual statement (e.g. “A shark interviews a fish”) does not make commonsense.
Prior work in diversification for \ac{GCR} has injected external knowledge from a knowledge graph~\citep{yu:2022:diversifying,hwang:2023:knowledge}, retrieved diverse sentences from an external corpora~\citep{dimongen}), or use in-context learning to instruct an \ac{LLM}~\cite{zhang:2024:diversity} to elicit diverse outputs.
However, our goal in this paper is \emph{not to propose diversification methods} for \ac{GCR}, but to conduct \emph{a meta-evaluation of existing metrics} proposed in prior work for evaluating the diversity of \ac{GCR}.

% We could put quality metrics here
\paragraph{Evaluating Quality in GCR:} 
Quality metrics in \ac{GCR} primarily assess coherence, logical consistency, and their correlation with human judgments~\citep{sai:2022:survey, yu:2022:diversifying}. 
Popular metrics use $n$-gram overlaps (e.g. BLEU~\citep{bleu}, ROUGE~\citep{rouge}), which measure the lexical overlap between a generated text and a human-written reference. 
BLEU~\citep{bleu}, for instance, computes the mean $n$-gram precision of a candidate sentence against human-written references, while semantic metrics (e.g. SPICE~\citep{spice}, BERTScore~\citep{zhang:2020:bertscore}) capture semantic textual similarity. BERTScore~\citep{zhang:2020:bertscore} uses contextualised word embeddings to measure the semantic overlap between tokens in paired sentences. 
Despite their wide use, quality metrics alone are insufficient for evaluating \ac{NLG} tasks, especially in \ac{GCR}.

\paragraph{Evaluating Diversity Metrics:}
Our work builds upon studies such as \citet{tevet2021evaluating}, who used human annotations to assess diversity metrics in \ac{NLG}.
 There are several important distinctions between their work and ours:
 \begin{enumerate}
     \item \textbf{Task-specific Focus:} They did not consider commonsense relations in the outputs they evaluate, which is an important requirement for \ac{GCR}.
     
     \item \textbf{Generation Variability:} They require adjustable decoding parameters (e.g. temperature) to control diversity.
    However, \citet{zhang:2024:diversity} showed that simply increasing temperature can harm the quality of commonsense generation.
    Instead, we use controlled perturbations (e.g. random shuffling and LLM-based paraphrasing) to generate outputs with varying diversity.

    \item \textbf{Annotation Methodology:} Whereas \citet{tevet2021evaluating} relied on crowdsourced human annotators---faced with low agreement and high cost---we leverage \acp{LLM} as reference-free annotators~\citep{wang:2023:chatgpt, geval, fu:2024:gptscore}. 
    Recent studies have successfully used \acp{LLM} for \ac{NLG} evaluations~\citep{kocmi:2023:large,geval} and \citet{zhang:2024:diversity} reported a moderate level of agreement between human and \ac{LLM}-based diversity ratings in \ac{GCR}.
    Our own human evaluation confirms that LLM-based diversity ratings achieve 80.6\% accuracy with expert human annotators.    
 \end{enumerate}
 In summary, while there has been extensive work on diversifying \ac{NLG} outputs and evaluating quality in \ac{GCR}, the evaluation of diversity metrics—especially in the context of commonsense generation—remains underexplored. Our work fills this gap by providing a systematic meta-evaluation of both form-based and content-based diversity metrics in GCR.

\section{Diversity Metrics for GCR}
\label{sec:diversity-metrics}

In this section, we describe the diversity metrics used in our meta-evaluation.

\paragraph{Form-based Diversity:}  
Self-BLEU~\citep{zhu:2018:selfbleu} measures the average $n$-gram overlap between all pairs of sentences within a set.\footnote{We subtract self-BLEU scores by $1$,  such that higher scores indicate greater pairwise diversity.}
We use self-BLEU-3/4 (i.e. $n=3, 4$) in our experiments. 
Inspired by ecology and quantum mechanics, \ac{VS} \citep{vendi:2022} was proposed as a diversity metric in computer vision.
\ac{VS} is the exponential of the Shannon's entropy over the eigenvalues of the pairwise similarity (kernel) matrix of a set of sentences, computed using either the $n$-gram overlap or sentence embeddings (see~\autoref{sec:vs_scores} for further details.)
\citet{cousins:2024} extended the original \ac{VS} by introducing an order parameter $q$, which adjusts its sensitivity to the frequency of the items.  
%In our experiments, we use VS with $q = 0.5/1/\infty$. 
A smaller $q$ (e.g. $q=0.5$)  increases the sensitivity to larger variances, capturing diversity more effectively in imbalanced scenarios, while $q=\infty$ is more robust against the intraclass variance, focusing on the most dominant features. 
For the form-based diversity measurement using \ac{VS}, the kernel matrix is constructed using a bag-of-$n$ grams representation.

Distinct-$k$~\citep{li:2016:distinct} calculates the ratio of the unique $k$-grams to the total number of $k$-grams, and is one of the widely-used metrics for evaluating corpus diversity. 
It adjusts the bias towards generating longer sequences, ensuring that diversity is not artificially inflated by the sentence length. 
Similarly, Entropy-$k$ quantifies the uniformity of the $k$-gram distribution within the text. 
Higher values for both Distinct-$k$ and Entropy-$k$ reflect greater diversity.

\paragraph{Content-based Diversity:}
To measure diversity at content level, self-CosSim~\citep{cox2021directed} calculates the average pairwise cosine similarity between the generated sentences using their sentence embeddings.
On the other hand, Chamfer Distance~\citep{chamfer:2006} measures diversity by calculating the average of the minimum pairwise distances between embeddings, reflecting proximity to the nearest neighbour (see~\autoref{sec:chamfer_dis}).
We also use \ac{VS} for content-based diversity, where the kernel matrix is built from sentence embeddings. 
For consistency across metrics, we use embeddings obtained via SimCSE~\citep{gao:2021:simcse}.

\section{Meta-Evaluation of Diversity Metrics}
\label{sec:methods}
We propose an \ac{LLM}-based annotation method for creating a diversity rated dataset for our meta-evaluation in \autoref{sec:llm_evaluator}, and a method to create sentence sets with different quality levels from the CommonGen dataset in \autoref{sec:candidate_sets}. Then we conduct a human evaluation on our \ac{LLM}-based diversity annotation in \autoref{sec:human_evaluation}.

\subsection{LLM-based Diversity Annotation}
\label{sec:llm_evaluator}

A reliable diversity metric must align well with the human notion of diversity, independently of the quality of the generation.
For example, randomly permuting the word order or including nonsensical words in a sentence are not considered by humans to be improving diversity.
Therefore, a reliable diversity metric must also not assign high diversity scores for such cases.
However, obtaining reliable human diversity ratings at scale is costly. 
Moreover, \citet{tevet2021evaluating} showed that human diversity judgments often conflate text quality and variety. 
Consequently, to conduct a large-scale meta-evaluation over existing diversity metrics, we elicit diversity ratings from an \ac{LLM}.
\acp{LLM} have been used as annotators for multiple \ac{NLG} tasks~\citep{wang:2023:chatgpt, geval, fu:2024:gptscore}.
In particular, \citet{zhang:2024:diversity} reported a moderate level of agreement between \ac{LLM} and human diversity ratings in a \ac{GCR} task.
% Crucially, \ac{LLM}-based diversity annotation allows us to separate out diversity from other confounding factors such as the overall quality, fluency and factuality of the generation.

We consider two types of diversities~\citep{tevet2021evaluating} in our annotation:
\paragraph{Form-based Diversity:}  A diverse set of sentences must exhibit minimal lexical overlap, avoiding repetitive word usage while preserving clarity and fluency.
\paragraph{Content-based Diversity:} A diverse set of sentences must exhibit distinct semantic content \emph{centred on the same input}, ensuring that each sentence offers a different perspective on the topic rather than talking about unrelated topics.

We ensure the quality of LLM diversity annotation through two steps:
\subsubsection{Prompt Engineering}
The prompt that we use to obtain diversity ratings from the LLM is shown in~\autoref{fig:llm_diversity_template} in the Appendix.
This prompt instructs the LLM diversity annotator to adhere to commonsense constraints (i.e. nonsensical outputs should not be interpreted to be genuinely diverse). 

We did not ask the annotator LLM to select a preferred set directly because the LLM exhibits ordering sensitivity problem that the ordering of choices would affect the quality ranking of candidates~\citep{wang:2024large,pezeshkpour:2024}. 
Our preliminary experiments confirmed that when we simply prompted the LLM diversity annotator to select the more diverse set, it chose Set 2 in 87.0 \% of sentence-set pairs on the CommonGen~\c test dataset with the generated sets, even after the sentence-set pairs were randomly swapped.

Therefore, we adopt the score-based method and instruct the annotator LLM to score each set’s diversity according to a five-point scale, from \emph{highly redundant} (1) to \emph{explore a wide range of aspects of the theme} (5). 
We also require that the annotator LLM consider thematic coherence among the sentences in a given set, when evaluating for their diversity. 
For each pair of candidate sentence sets, we prompt the annotator LLM five times and average the predicted diversity ratings per sentence-set and determine the set with the higher mean rating as the more diverse one.

We use \texttt{GPT-4o} as the annotator \ac{LLM}, which has shown superior performance in a broad range of annotation tasks.\footnote{\href{https://artificialanalysis.ai/leaderboards/models}{LLM Leaderboard}}
Prior work using LLMs for rating NLG tasks have shown that \texttt{GPT-4o} to demonstrate stronger correlations with human ratings~\cite{geval,bai2024benchmarking}. 

\subsubsection{Few-shot Prompting}
\begin{figure*}[t]
\centering
\includegraphics[width=1.0\linewidth]{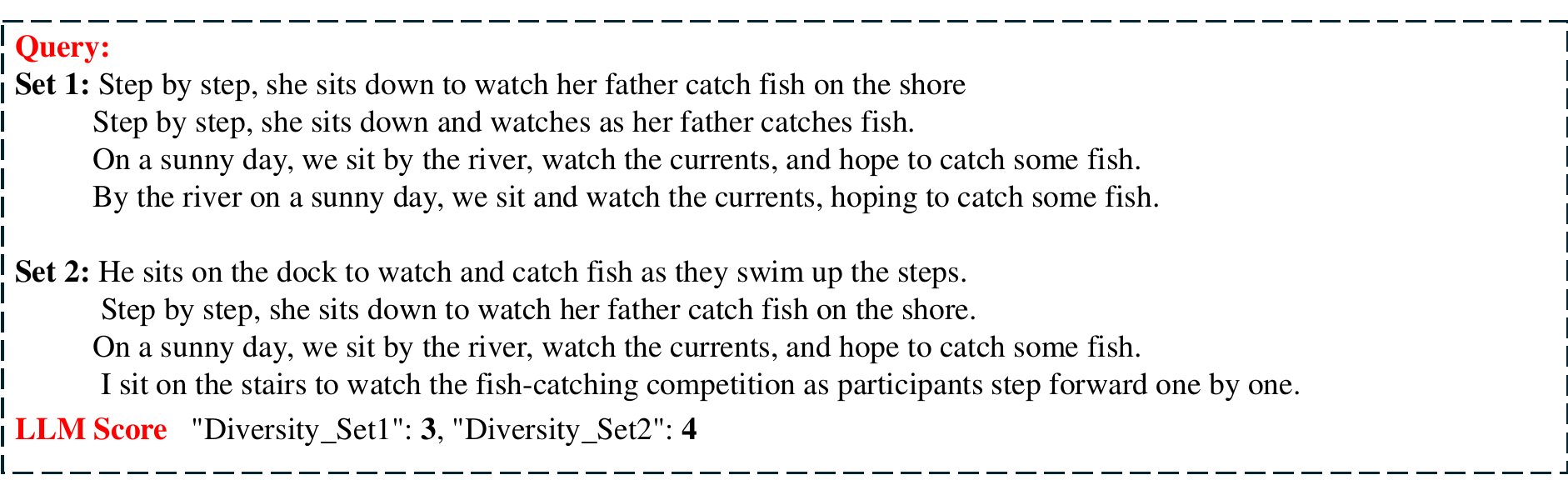}
\caption{An example of annotating for diversity using \texttt{GPT-4o} for two sets of sentences generated for the same input concepts. \texttt{GPT-4o} assigns a higher diversity rating for Set-2, indicating it to be more diverse than Set-1.}
\label{fig:llm_reason_example}
\end{figure*}

In-context Learning (ICL) has proven to be an effective strategy for improving text generation and evaluation in many \ac{NLG} tasks~\citep{GPT3:2020,dong:2022:icl}.
Diversity evaluation presents significant challenges for LLM alignment with human judgments.
Consequently, to guide \texttt{GPT-4o} towards human-like diversity judgments, we create a set of human-labelled few-shot examples illustrating how diverse (or non-diverse) outputs should be scored.
Specifically, we asked three linguistically trained annotators to independently score the diversity of 60 sentence sets. 
Each set comprises of four sentences generated by the same model from the same input concepts.
Specifically, each annotator is instructed to:
\begin{enumerate}
    \item Assign a 1--5 rating to each sentence set.
    \item Rank the sets (if they shared the same input) with their diversity preference. This ranking resolves ties when two sets receive the same numerical score.
\end{enumerate}
Finally, we select the top 8 sentence set pairs with the highest agreement among the human annotators as the few-shot examples to be included in our prompt.
An example of an LLM-based diversity judgement by \texttt{GPT-4o} is shown in~\autoref{fig:llm_reason_example}.

\subsection{Candidate Sets}
\label{sec:candidate_sets}

Diversity would be of interest only when the generation quality is high. 
% It would not be meaningful to evaluate diversity in a poor quality generation.
Therefore, a reliable diversity metric must be able to accurately evaluate the diversity of generations of varying qualities. 
For this purpose, we propose a method to create sentence sets that have varying levels of generation quality to be used later in our meta-evaluations.
Specifically, we use the CommonGen dataset~\citep{CommonGen} where a \ac{GCR} model must generate a coherent sentence that contains all of the input concepts, reflecting their commonsense relations.
We use the official CommonGen test set, which includes 1,497 examples, each containing 3--5 input concepts on average.
We create sets of sentences of \emph{high} and \emph{low} generation quality as described respectively in \autoref{sec:high_quality_sets} and \autoref{sec:low_quality_sets} by prompting three \textbf{generator LLMs}\footnote{To prevent any confusion with the \texttt{GPT-4o} that we used as the annotator LLM in \autoref{sec:llm_evaluator}, we collectively call those models as the \textbf{generator LLMs}.}: \texttt{GPT-4-turbo}~\citep{gpt4}, \href{https://huggingface.co/meta-llama/Llama-3.1-8B-Instruct}{\texttt{Llama3.1-8b}}~\citep{llama3.1}, and \href{https://huggingface.co/Qwen/Qwen2.5-14B-Instruct}{\texttt{Qwen 2.5-14b}}~\citep{qwen2.5}. 
Due to space limitations, we show the detailed instructions provided to the generator LLMs, an empirical quality evaluation, and example generations in \autoref{sec:app:generator-LLMs}.

\subsubsection{High-Quality Sentence Sets} 
\label{sec:high_quality_sets}

We propose the following strategies to create sentence sets with high generation quality.
\paragraph{Default:}
Note that CommonGen was developed as a dataset for evaluating the quality and not diversity of \ac{GCR} methods.
Therefore, it contains only a small number of human-written sentences covering the input concepts in a test case. 
Moreover, these human-written sentences do not adequately cover all possible commonsense bearing sentences that can be generated from the input concepts.
To address this issue, we prompt the generator LLMs with the same instructions as given to the human annotators in CommonGen to generate four sentences for each test case, as four is the average number of sentences per test instance in the CommonGen dataset.
%Typically, it is easier to increase the diversity in a set by generating more sentences, while it is more challenging to be diverse within a smaller set, 
This enables us to evaluate the reliability of diversity measures more accurately.
We call it the \textbf{Default} set of sentences for a test case. 

\paragraph{Paraphrasing:}
We randomly select one or more sentences from the \textbf{Default} set and instruct the generator LLMs to create their paraphrases.
We then replace the non-selected sentences in each Default set with the generated paraphrase sentences.
We expect the diversity of a set of sentences to decrease when we include more paraphrasing sentences.
Specifically, we consider three variants of this method.
Let the Default set contain four sentences $\{A, B, C, D\}$, and a $A^*$ be the paraphrase of $A$, selected randomly from the set.
We then define: \textbf{Para-1} = $\{A, A^*, B, C\}$, \textbf{Para-2} = $\{A, A^*, B, B^*\}$, and \textbf{Para-3} = $\{A, A^*, A^{**}, B\}$.

\subsubsection{Low-Quality Sentence Sets} 
\label{sec:low_quality_sets} 

To evaluate the ability of a diversity metric to accurately distinguish genuine diversity from nonsensical or random corruptions made to a sentence, we create a set of low generation quality sentences for each input concept set in CommonGen test dataset as follows. 
\paragraph{Nonsensical:}
We prompt\footnote{The specific prompt is shown in \autoref{sec:app:generator-LLMs}.} the generator LLMs to produce sentences that are syntactically valid and include all of the input concepts, but do not make any commonsense or illogical.

\paragraph{NounShuff:}
We run a part-of-speech tagger and randomly shuffle nouns and pronouns within each sentence, while leaving other words unchanged. 
This process disrupts semantic consistency while retaining some semblance of syntactic framing, serving as an intermediate case of corruption.

\paragraph{RndShuff:}
We take each sentence from the \textbf{Default} set and randomly shuffle \emph{all of the words} in it to produce sequences that are devoid of coherent sentence structure or meaning.

\subsection{Human Evaluation of the LLM Annotator}
\label{sec:human_evaluation}
% Doing human evaluation on the LLM
To assess the reliability of our \ac{LLM}-based diversity annotations, we randomly selected 70 pairs of high-quality sentence sets and asked five linguistically trained human annotators (graduate students and academics trained in linguistic annotation tasks) to indicate which set they judged to be more diverse.
Unlike the \ac{LLM}, which can be sensitive to the input set ordering, human annotators are not susceptible to the ordering of candidate set pairs. We therefore showed both candidate sets, using the same diversity criteria provided to the \ac{LLM}, but asked the human annotators to choose their preferred set directly, without rating each set individually.
We then compared the human annotations with the \ac{LLM}’s preferences.

To measure the agreement, we calculated the pairwise accuracy between each human annotator’s judgments and the \ac{LLM} annotator's decisions for all pairs. 
The average pairwise accuracy across all annotators was then computed to represent the overall agreement. 
The resulting agreement of 80.6\% demonstrates that our \ac{LLM}-based annotations provide an accurate and reliable alternative to human diversity judgments.

\citet{tevet2021evaluating} highlighted that evaluating text diversity is challenging for crowdsourced human annotators, as judgments can be influenced by individual biases or lack of linguistic training. 
We calculated Fleiss’ Kappa to measure agreement among the five human annotators, which was
 0.45 and indicates a moderate level of agreement among the human annotators, demonstrating the difficulty of this annotation task.

\section{Experiments}
\label{sec:exp}

\subsection{Settings and Evaluation Metrics}
\label{sec:exp:settings}

To obtain statistically stable diversity ratings, we run the annotator LLM (i.e. \texttt{GPT-4o}) with the \emph{temperature} set to $1.0$ (further details on temperature tuning are provided in~\autoref{sec:temperature_tuning}), and average the results over five independent runs.
All experiments are conducted on two GPUs  (Nvidia A6000 and 4090) for the \texttt{Qwen2.5-14B} and \texttt{Llama3.1-8B} models.
For \texttt{GPT-4-turbo}, we use the OpenAI API, with the temperature set to $0$ to increase determinism in the generations.
We use 1024-dimensional\footnote{\url{huggingface.co/princeton-nlp/sup-simcse-roberta-large}} SimCSE~\cite{gao:2021:simcse} sentence embeddings for all content-based diversity metrics.

We define the \emph{accuracy} of a target diversity metric as the percentage of pairwise decisions that agree with those of the annotator \ac{LLM}.
For example, given a pair of sentence sets $(\cS_1, \cS_2)$, if both the annotator LLM and the target diversity metric consider $\cS_1$ to be more diverse than $\cS_2$, it is counted as a correct prediction.
To prevent diversity evaluations from being influenced by the quality of the sentence sets, we ensure that both sentence sets in a pair to have the same generation quality (i.e. both sets must be of either high quality or low quality).
Moreover, to ensure meaningful comparisons, we filter out any sentence set pairs where the annotator LLM's average diversity ratings differ by less than 0.5. 
After this filtering step, the resulting sentence pair sets generated with \texttt{GPT-4-turbo}, \texttt{Llama3.1-8B}, and \texttt{Qwen-2.5-14B} and used for evaluations contain, respectively, 1414, 1916, and 1864 instances.

\begin{table}[t]
\centering
\resizebox{\columnwidth}{!}{
\begin{tabular}{clccc}
\toprule
 & \textbf{Diversity Metric}        & \textbf{GPT-4-turbo}  & \textbf{Qwen2.5-14B}  & \textbf{Llama3.1-8B}  \\
\midrule
\multirow{7}{*}{\rotatebox{90}{Form}} & self-BLEU-3                        & 48.4\textsubscript{±2.60}                           & 50.7\textsubscript{±2.26}                             & 52.7\textsubscript{±2.24}                         \\
& self-BLEU-4                        & 49.0\textsubscript{±2.60}                           & 51.9\textsubscript{±2.27}                             & 53.0\textsubscript{±2.23}                         \\
& \ac{VS}-ngram-0.5                    & 49.2\textsubscript{±2.61}                           & 57.7\textsubscript{±2.24}                             & 56.1\textsubscript{±2.22}                         \\
& \ac{VS}-ngram-1                      & 49.0\textsubscript{±2.60}                           & 57.8\textsubscript{±2.24}                             & 56.2\textsubscript{±2.22}                         \\
& \ac{VS}-ngram-inf                    & 47.5 \textsubscript{±2.60}                          & 58.9\textsubscript{±2.23}                             & 56.5\textsubscript{±2.22}                         \\
& Distinct-4                         & 64.0\textsubscript{±2.49}                           & 69.0\textsubscript{±2.09}                             & 61.7\textsubscript{±2.18}                         \\
& Entropy-2                          & 62.9\textsubscript{±2.52}                           & 74.0\textsubscript{±1.99}                             & 62.5\textsubscript{±2.17}                         \\
\midrule
\multirow{5}{*}{\rotatebox{90}{Content}} & Chamfer                             & 80.6\textsubscript{±2.06}                           & 78.9\textsubscript{±1.85}                             & 71.9\textsubscript{±2.01}                         \\
& self-cosSim                         & 76.9\textsubscript{±2.20}                           & 80.0\textsubscript{±1.81}                             & 71.9\textsubscript{±2.01}                         \\
& \ac{VS}-Embed-0.5                    & 80.7\textsubscript{±2.06}                         & 80.8\textsubscript{±1.79}                             & 73.2\textsubscript{±1.98}                         \\
& \ac{VS}-Embed-1                      & 79.3\textsubscript{±2.11}                           & 81.1\textsubscript{±1.78}                             & 73.1\textsubscript{±1.99}                         \\
& \ac{VS}-Embed-inf                    & 76.0\textsubscript{±2.21}                           & 79.9\textsubscript{±1.81}                             & 71.9\textsubscript{±2.01}                         \\
\bottomrule
\end{tabular}
}
\caption{Meta-evaluation of the accuracy of the diversity metrics on the CommonGen test dataset with each of the generator LLMs, with 95\% bootstrap CI half-widths in subscripts.}
\label{tab:diversity_metrics_models}
\end{table}

\subsection{Meta-Evaluation of Diversity Metrics}
\label{sec:meta-eval}

\autoref{tab:diversity_metrics_models} shows the accuracy of form-based (top group) vs. content-based (bottom group) \ac{GCR} diversity metrics on the CommonGen dataset.
We observe that content-based diversity metrics-—-specifically self-cosSim, Chamfer, and \ac{VS}-Embed variants---consistently achieve higher accuracy than form-based diversity metrics such as the corpus-level diversity metrics (e.g. Entropy, Distinct) or the $n$-gram-based diversity metrics (e.g. self-BLEU, \ac{VS}-$n$-gram variants) across all generator LLM outputs.
In particular, \ac{VS}-Embed-0.5 and \ac{VS}-Embed-1 consistently report the best accuracy, suggesting that the content is more important than the form when evaluating diversity in \ac{GCR}.
Form-based metrics primarily focus on lexical overlap, overlooking the deeper semantic nuances that characterise the diversity. 
Although Entropy and Distinct reflect some aspects of overall lexical variety and frequency distributions, they fail to capture semantic richness. 
Even when these metrics sometimes outperform self-BLEU, they still fall short of content-based metrics.

\begin{table}[t]
\centering
\resizebox{0.8\columnwidth}{!}{
\begin{tabular}{clcc}
\toprule
 & \textbf{Diversity Metric} & \textbf{ComVE} & \textbf{DimonGen} \\
\midrule
\multirow{7}{*}{\rotatebox{90}{Form}} 
    & self-BLEU-3        & 77.3\textsubscript{±2.76}  & 59.7\textsubscript{±3.18}  \\
    & self-BLEU-4        & 76.9\textsubscript{±2.78}  & 59.4\textsubscript{±3.19}  \\
    & \ac{VS}-ngram-0.5   & 76.7\textsubscript{±2.78}  & 60.0\textsubscript{±3.18}  \\
    & \ac{VS}-ngram-1     & 77.0\textsubscript{±2.77}  & 59.8\textsubscript{±3.18}  \\
    & \ac{VS}-ngram-inf   & 77.2\textsubscript{±2.76}  & 58.8\textsubscript{±3.19}  \\
    & Distinct-4         & 73.8\textsubscript{±2.90}  & 62.2\textsubscript{±3.14}  \\
    & Entropy-2          & 74.2\textsubscript{±2.89}  & 62.2\textsubscript{±3.14}  \\
\midrule
\multirow{5}{*}{\rotatebox{90}{Content}} 
    & Chamfer            & 77.0\textsubscript{±2.73}  & 67.8\textsubscript{±3.03} \\
    & self-cosSim        & 76.4\textsubscript{±2.80}  & 66.6\textsubscript{±3.06}  \\
    & \ac{VS}-Embed-0.5   & 77.4\textsubscript{±2.76}  & 67.2\textsubscript{±3.04}  \\
    & \ac{VS}-Embed-1     & 76.8\textsubscript{±2.78}  & 67.6\textsubscript{±3.04}  \\
    & \ac{VS}-Embed-inf   & 76.4\textsubscript{±2.80}  & 66.6\textsubscript{±3.06}  \\
\bottomrule
\end{tabular}
}
\caption{Accuracy of diversity metrics on ComVE and DimonGen datasets, with 95\% bootstrap CI half-widths in subscripts.}
\label{tab:diversity_metrics_comve_dimongen}
\end{table}

% Kappa coefficient
\begin{figure*}[t]
\centering
\includegraphics[width=1.0\linewidth]{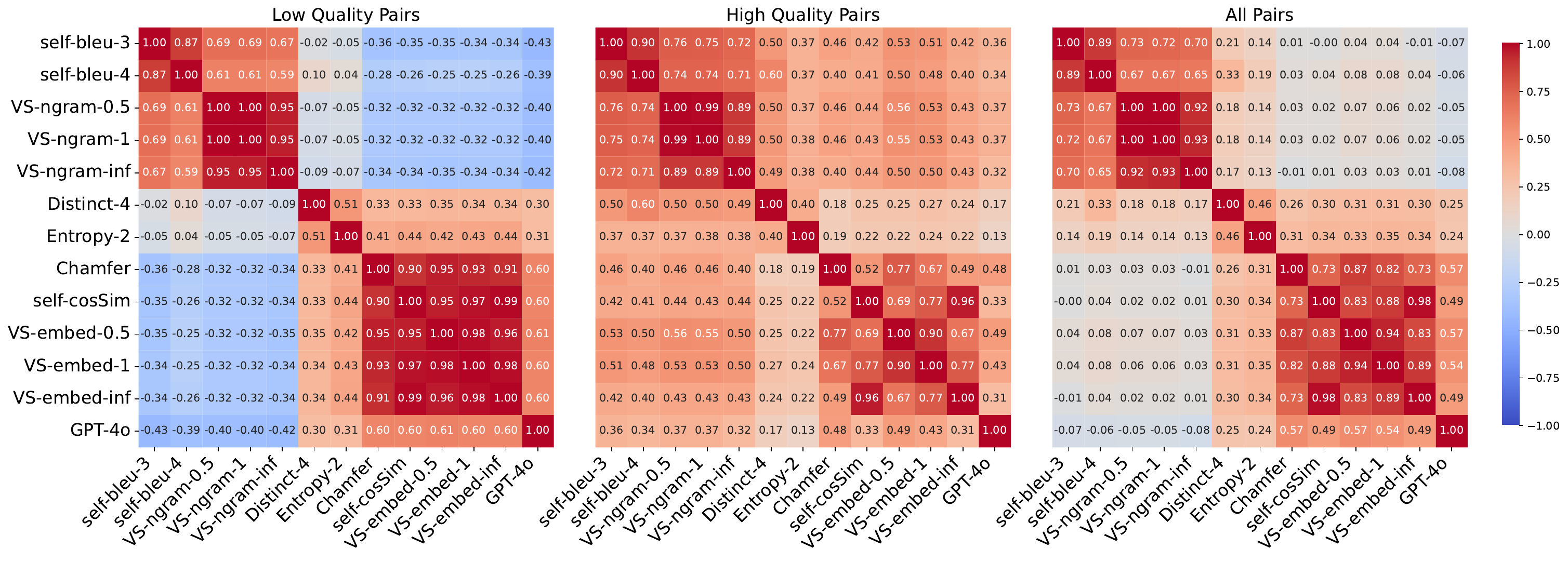}
\caption{Inter-annotator agreement (measured using Cohen's Kappa) between two diversity metrics when used to make pairwise preference orderings for sentence sets generated for the same input concepts in CommonGen test cases. Agreement with the annotator LLM (i.e. \texttt{GPT-4o}) is also shown.}
\label{fig:kappa_combined}
\end{figure*}

To ensure that our findings generalise beyond CommonGen, we extend the meta-evaluation to two additional commonsense generation datasets: ComVE~\citep{semeval} and DimonGen~\citep{dimongen}. 
ComVE requires a \ac{GCR} method to explain why a counterfactual statement is nonsensical, while DimonGen focuses on generating diverse sentences describing relationships between two given concepts. 
Both tasks require outputs that are diverse and commonsense-bearing.
\citet{zhang:2024:diversity} provide three sets of generated sentences for each dataset, along with a pre-evaluation of output quality. 
We compare each pair of sentence sets generated for the same input using the diversity ratings returned by our annotator LLM (i.e. \texttt{GPT-4o}), and contrast these with the diversity scores produced by each target metric, as shown in \autoref{tab:diversity_metrics_comve_dimongen}.

Consistent with the trends observed on CommonGen, \textbf{content-based metrics} (e.g. \ac{VS}-Embed-0.5,  Chamfer) consistently achieve the highest agreement with \texttt{GPT-4o} on both ComVE and DimonGen.
For example, \ac{VS}-Embed-0.5 performs best on ComVE, whereas Chamfer excels on DimonGen. 
Although form-based metrics show competitive accuracies on the ComVE dataset, their performance drops on DimonGen.  
These findings confirm that content-based metrics offer a more reliable and consistent approach for evaluating text diversity, especially in diverse commonsense generation tasks.
While form-based metrics have close alignment with content-based metrics on the ComVE dataset, their performance is not always consistent (see~\autoref{sec:further_comve}).

\subsection{Diversity Metrics and Generation Quality}
\label{sec:high-vs-low-eval}

\begin{table}[t]
\centering
\resizebox{\columnwidth}{!}{
\begin{tabular}{c l|cc|cc|cc}
\toprule
 &  & \multicolumn{2}{c|}{\textbf{GPT-4-turbo}} & \multicolumn{2}{c|}{\textbf{Qwen2.5-14b}} & \multicolumn{2}{c}{\textbf{Llama3.1-8b}} \\
\cmidrule(lr){3-4} \cmidrule(lr){5-6} \cmidrule(lr){7-8}
 & \textbf{Diversity Metric} & \textbf{High} & \textbf{Low} & \textbf{High} & \textbf{Low} & \textbf{High} & \textbf{Low} \\
\midrule
\multirow{7}{*}{\rotatebox{90}{Form}} 
    & self-BLEU-3       & 73.5 & 27.6 & 68.4 & 35.3 & 66.6 & 39.8  \\
    & self-BLEU-4       & 72.0 & 30.0 & 67.1 & 38.7 & 64.3 & 42.5  \\
    & VS-ngram-0.5      & 73.7 & 28.8 & 69.7 & 47.2 & 66.5 & 46.5  \\
    & VS-ngram-1        & 73.4 & 28.8 & 69.5 & 47.6 & 66.6 & 46.8  \\
    & VS-ngram-inf      & 71.0 & 27.8 & 69.7 & 48.0 & 67.0 & 46.8  \\
    & Distinct-4        & 61.7 & 65.9 & 58.6 & 79.4 & 56.3 & 66.6  \\
    & Entropy-2         & 59.2 & 65.9 & 57.0 & 88.5 & 49.6 & \textbf{74.4}  \\
\midrule
\multirow{5}{*}{\rotatebox{90}{Content}} 
    & Chamfer           & \textbf{80.2} & 80.8 & 67.5 & \textbf{88.9} & 73.6 & 70.4  \\
    & self-cosSim       & 72.3 & 80.7 & 71.7 & 87.2 & 74.4 & 69.6  \\
    & VS-Embed-0.5      & \textbf{80.2} & \textbf{81.1} & 72.3 & 88.2 & \textbf{76.9} & 69.8  \\
    & VS-Embed-1        & 77.7 & 80.6 & \textbf{73.0} & 88.1 & \textbf{76.9} & 69.5  \\
    & VS-Embed-inf      & 71.2 & 80.7 & 71.5 & 87.3 & 74.4 & 69.6  \\
\bottomrule
\end{tabular}
}
\caption{Accuracy of diversity metrics across different levels of quality in sentence sets, generated by three generator LLMs. The content-based diversity metrics consistently perform better than the form-based metrics. Highest mean accuracy on each set is bolded.}
\label{tab:commongen_diversity_alignments}
%\vspace{-3mm}
\end{table}

\begin{table*}[t]
\centering
\resizebox{\textwidth}{!}{
\begin{tabular}{clccccccc}
\toprule
 &  & \multicolumn{4}{c}{\textbf{High-quality candidate sets}} & \multicolumn{3}{c}{\textbf{Low-quality candidate sets}} \\
\cmidrule(r){3-6} \cmidrule(l){7-9}
 & \textbf{Diversity Metric} & \textbf{Default} & \textbf{Para-1} & \textbf{Para-2} & \textbf{Para-3} & \textbf{Nonsensical} & \textbf{NounShuff} & \textbf{RndShuff} \\
\midrule
\multirow{7}{*}{\rotatebox{90}{Form}} 
 & self-BLEU-3            & 79.53 & 73.37 & 64.24 & 63.32 & 80.96 & 89.06 & 96.75 \\
 & self-BLEU-4            & 87.63 & 81.86 & 73.98 & 71.88 & 89.79 & 95.19 & 99.05 \\
 & VS-ngram-0.5   & 3.90  & 3.86  & 3.80  & 3.77  & 3.90  & 3.93  & 3.95  \\
 & VS-ngram-1     & 3.79  & 3.72  & 3.62  & 3.57  & 3.81  & 3.87  & 3.89  \\
 & VS-ngram-inf & 2.60  & 2.48  & 2.38  & 2.26  & 2.60  & 2.77  & 2.84  \\
 & Distinct-4             & 93.04 & 90.60 & 90.93 & 89.90 & 90.26 & 98.06 & 99.94 \\
 & Entropy-2              & 9.52  & 9.42  & 9.60  & 9.66  & 9.12  & 9.82  & 10.23 \\
\midrule
\multirow{5}{*}{\rotatebox{90}{Content}} 
 & self-CosSim            & 26.81 & 22.04 & 20.03 & 17.50 & 42.02 & 28.83 & 27.57 \\
 & Chamfer                & 20.09 & 12.44 &  3.09 &  9.54 & 35.18 & 22.53 & 21.34 \\
 & VS-Embed-0.5  & 2.67  & 2.35  & 2.08  & 2.08  & 2.59  & 2.77  & 2.72  \\
 & VS-Embed-1    & 2.01  & 1.76  & 1.60  & 1.55  & 2.63  & 2.09  & 2.04  \\
 & VS-Embed-inf & 1.26  & 1.20  & 1.18  & 1.15  & 2.52  & 1.28  & 1.27  \\
\bottomrule
\end{tabular}
}
\caption{Average diversity score of each metric on sentence sets generated using the methods described in \autoref{sec:candidate_sets}. For the high quality candidates, the diversity decreases from the \textbf{Default} set to \textbf{Para-3} set. Meanwhile, the low-quality sets are assigned with higher diversity scores.}
\label{tbl:operators}
\end{table*}

\begin{figure*}[t]
\centering
\begin{subfigure}{0.49\linewidth}
\centering
\includegraphics[width=\linewidth]{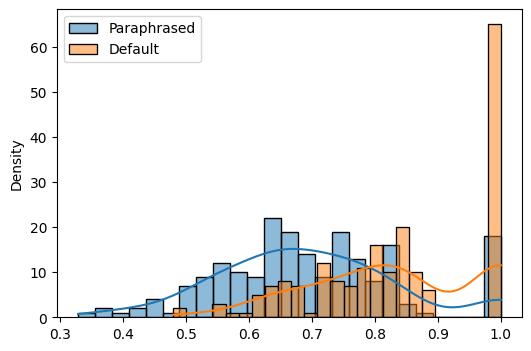}
\subcaption{Distribution of self-BLEU-3 scores}
\label{fig:histo_sb3}
\end{subfigure}
\begin{subfigure}{0.49\linewidth}
\centering
\includegraphics[width=0.96\linewidth]{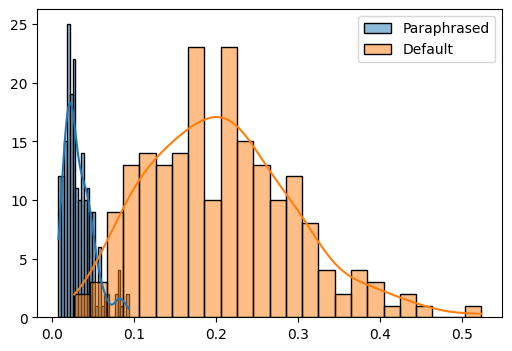}
\subcaption{Distribution of Chamfer scores}
\label{fig:histo_chamfer}
\end{subfigure}
\caption{Distribution of diversity scores for self-BLEU-3 (form-based) and Chamfer Distance (content-based) for \textbf{Default} and \textbf{Paraphrased} high-quality sentence sets. In self-BLEU-3, the two distributions have a high overlap, whereas in Chamfer they are well-separated. This indicates that the Chamfer metric can better distinguish more diverse \textbf{Default} sentence sets from the less diverse \textbf{Paraphrased} sentence sets than self-BLEU-3.}
\label{fig:histo_combined}
\end{figure*}

In \autoref{tab:commongen_diversity_alignments}, we conduct a meta-evaluation of diversity metrics for their ability to reliably estimate diversity in both high and low quality generations.
We see that form-based metrics perform particularly well when the generation quality is high, however, their accuracy drops drastically (even below 40\%) for low quality sets, demonstrating their sensitivity to inherent noise in the $n$-gram overlaps.
In contrast, content-based metrics maintain consistently high accuracy, regardless of generation quality.  
In particular, \ac{VS}-Embed-0.5 and \ac{VS}-Embed-1 approach or exceed 70\% accuracy in all comparisons, even for shuffled or nonsensical scenarios, demonstrating statistically significant improvements (see \autoref{sec:app:CI} for more details) over form-based metrics.

We treat each diversity metric as an \emph{annotator} that provides a preference ordering for diversity between two sentence sets, and measure their pairwise agreements.
We use Cohen's Kappa (shown in \autoref{fig:kappa_combined}) for this purpose, which is known to be less sensitive to class imbalance, and more reflective of true, non-random agreement. 
For high quality sets, most metrics achieve fair to substantial levels of agreement, reflecting strong consistency.
However, agreements vary considerably in low quality sets.
Content-based metrics such as Chamfer, self-cosSim, and \ac{VS}-Embed variants exhibit near-perfect agreement with each other and maintain Kappa values exceeding 0.6 with \texttt{GPT-4o}.
Conversely, form-based metrics (e.g. self-BLEU) show poor agreement with \texttt{GPT-4o} in low quality sets with negative Kappa values indicating that the observed agreement between these form-based metrics is lower than would be expected by chance.
Moreover, the agreements between form- and content-based metrics remain low, underscoring fundamental differences in how these metrics measure diversity.
Notably, Distinct-4 and Entropy-2---although also use $n$-grams---are less likely to overemphasise repeated phrases or minor word swaps and show a moderate level of agreement with content-based metrics even for low quality sets.

\autoref{tbl:operators} shows the average diversity score reported by each metric over the sets of sentences generated from \texttt{GPT-4-turbo} according to the high and low quality preserving methods described in \autoref{sec:candidate_sets}.
For the high quality candidates, as expected, we see that the diversity decreases from the \textbf{Default} set as we paraphrase more sentences (\textbf{Para-1} to \textbf{Para-3}), as measured by all metrics.
We also find that, on average, all metrics assign higher diversity scores to low quality generations than to high quality generations. 
This is because a random set of sentences could appear to be diverse, covering distinct topics, at both the form and content.
This observation highlights an important limitation of existing \ac{GCR} diversity evaluation metrics: diversity should \emph{not} be evaluated without considering quality.
A promising future research direction would be to develop an evaluation metric for \ac{GCR} that simultaneously incorporates both quality and diversity aspects.

% TZ: A new draft of to discribe figure 4
% The distribution of diversity scores assigned by self-BLUE3 (form-based) and Chamfer (content-based) for high-quality (Default) and low-quality (Nonsensical) candidate sets are shown in \autoref{fig:histo_combined}. 
% Self-BLEU-3 (~\autoref{fig:histo_sb3}) exhibits a highly skewed distribution, with most scores concentrated near the upper bound (close to 1). Furthermore, the curves for High-quality and Low-quality sets are nearly overlapping, indicating that self-BLEU3  is sensitive to word overlap  and fails to capture meaningful diversity. 
% In contrast, Chamfer (~\autoref{fig:histo_chamfer}) demonstrates a wider score distribution and better separation between High-quality and Low-quality sets. 
% While the average Chamfer score is higher for low-quality sets, this does not imply that Chamfer prefers low-quality outputs. Instead, it shows the ability of Chamfer to distinguish between sentence sets of different quality levels and its score distributions for high- and low-quality sets remain well-separated, indicating that it captures genuine semantic distinctions. Additionally, our evaluation does not directly compare raw Chamfer scores across quality levels; instead, we assess how well each metric aligns with diversity preferences within matched-quality comparisons.

\autoref{fig:histo_combined} compares the distribution of diversity scores assigned by self-BLEU-3 (form-based) versus Chamfer (content-based) for 200 randomly sampled sentence sets from \textbf{Default} and \textbf{Paraphrased} (using \textbf{Para-2}) high-quality candidate sets. 
Sentence sets in \textbf{Paraphrased} are constructed to be less diverse compared to those in \textbf{Default}.
We use Kernel Density Estimation~\cite{KDE} to interpolate the distributions from the frequency histograms.
We see that the two distributions for self-BLUE-3 in \autoref{fig:histo_sb3} to have a high overlap, demonstrating its inability to correctly separate high diversity generations in \textbf{Default} from the less diverse generations in \textbf{Paraphrased}. 
On the other hand, the two distributions for Chamfer in \autoref{fig:histo_chamfer} exhibit a relatively smaller overlap, indicating that Chamfer assigns relatively higher diversity scores to the sentence sets in \textbf{Default} than those in \textbf{Paraphrased}.
%We  that self-BLEU-3 (\autoref{fig:histo_sb3}) exhibits a highly skewed distribution with the KDE curves under the self-BLEU-3 are extremely close, indicating that self-BLEU-3 is overly sensitive to surface word overlap and fails to distinguish between truly meaningful variations. 
%In contrast,  Chamfer (~\autoref{fig:histo_chamfer}) demonstrates a wider score distribution and better separation between the sets. It shows the ability of Chamfer to distinguish between sentence sets of different diversity levels.

\section{Conclusion}
We presented a comprehensive meta-evaluation of diversity metrics for commonsense generation, revealing that content-based metrics consistently align with human judgments while form-based metrics tend to overestimate diversity, especially in low-quality generations. 
Our experiments across multiple datasets demonstrate that metrics such as \ac{VS}-Embed and Chamfer provide a more robust and reliable assessment of semantic diversity. 
These findings underscore the importance of incorporating content-level analysis in evaluating commonsense generation. 
%Our work motivates the development of unified metrics that simultaneously account for both quality and diversity. 
Future research should build on these insights to further enhance the robustness and interpretability of \ac{GCR}.

\section{Limitations}

The experiments conducted in this paper were limited to English, a morphologically limited language.
Although we would like to extend our meta-evaluation to other languages, we were limited by the lack of availability of commonsense reasoning datasets for languages other than English.
In particular, CommonGen~\citep{CommonGen}, ComVE~\citep{semeval}, and DimonGen~\citep{dimongen} datasets are specifically designed for evaluating diversified commonsense reasoning only in English.
We note however that both form- and content-based diversity metrics considered in our work are not limited to English, and can be easily extended to other languages with suitable tokenisers or multilingual sentence embedding models.
For example, a single Kanji character in languages such as Japanese or Chinese can carry meaning on its own, and even $n$-gram overlap measures defined over character sequences can capture some level of meaning retention between a generated and a reference set of sentences.
Therefore, we believe it would be important to conduct similar meta-evaluation for the diversity metrics in commonsense generation for other languages before selecting an appropriate evaluation metric.
We hope that the methodology we propose in this paper will be exemplary in such future work.

Our work evaluates diversity metrics primarily within \ac{GCR} tasks. 
The candidate sets used in this study were pre-evaluated for quality using official scripts (for CommonGen) or prior work (for ComVE and DimonGen).
We use three \acp{LLM} as our generative models, a closed model (\texttt{GPT4-turbo}) and two open-source models (\texttt{Llama3.1-8B} and \texttt{Qwen2.5-14B}) to promote the reproducibility of our results, which are reported using multiple publicly available benchmarks.
Of course, there is a large number of LLMs being developed, trained on different pre-train data compositions, architectures, parameter sizes and fine-tuned for a plethora of tasks.
It is practically impossible to consider all available LLMs in a conference paper due to the sheer number and the computational costs.

We used \texttt{GPT-4o} as the sole \ac{LLM}-based diversity annotator.
Although the prompts and instructions are adaptable to other models, we chose \texttt{GPT-4o} due to its superior performance in a range of NLG tasks. 
Moreover, in our human evaluation, conducted over a subset of the \texttt{GPT-4o} rated sentence sets, human judges found those annotations to be of high accuracy (i.e. 80.6\% accuracy as shown in \autoref{sec:human_evaluation}).
Therefore, we consider \texttt{GPT-4o} to offer a scalable and robust alternative for annotating diversity in sentence sets.
However, using LLMs that are comparable or superior to \texttt{GPT-4o} could further validate our findings.

\section{Ethical Concerns}
All experiments conducted in this study use publicly available datasets, CommonGen, ComVE, and DimonGen.
To the best of our knowledge no personally identifiable information is included in those datasets and no ethical issues have been reported.
The human annotators who participated in our evaluation were over 18 years old adults and have given informed consent to use their diversity annotations for academic research purposes.

It is noteworthy that \acp{LLM} have been reported to encode social biases such as gender or racial biases~\cite{Kaneko:2021,CrowsPairs,Kaneko:2022a}.
Although we evaluated quality and diversity of the generations made by \acp{LLM} in this work, we have not evaluated how social biases are reflected in their generations.
Therefore, it is important to also evaluate the social biases in the diverse \ac{LLM} generations before a diversification method for \ac{GCR} is deployed in an \ac{NLG} application.

\bibliography{myrefs.bib}

\section*{Supplementary Materials}
\appendix

\section{Vendi Score (VS)}
\label{sec:vs_scores}
The \ac{VS} is a similarity-based diversity metric, inspired by the ecological diversity, which is defined as the exponential of the entropy of the distribution of the species under study. 
Specifically, \ac{VS} calculates the exponential of the Shannon entropy of the eigenvalues of a similarity matrix~\citep{vendi:2022}.
Let $\mathbf{K} \in \mathbb{R}^{n \times n}$ be the kernel matrix with entries $K_{i,j} = k(x_i, x_j)$. In our experiments, $k(x_i, x_j)$ is computed as the dot product of the $n$-gram (for form-based diversity evaluations) or pre-trained embedding (for content-based diversity evaluations) of each sentence pair in a set of sentences. 
Let us denote the eigenvalues of $\mathbf{K}$ by $\lambda_1, \lambda_2, \ldots, \lambda_n$. 
Then, \ac{VS} is given by \eqref{eq:vs}.
\begin{align}
    \label{eq:vs}
    VS = \exp\left(-\sum_{i=1}^n \lambda_i \log \lambda_i\right)
\end{align}
The \ac{VS} could be interpreted as the effective number of dissimilar elements in a sample. 
This formulation corresponds to a special case where the order $q=1$. 
However, it has the limitation that it could not handle imbalanced datasets where rare elements might be under-represented. 
To address these challenges, the \ac{VS} has been generalised to include different orders $q$~\citep{cousins:2024} as given by \eqref{eq:vs_cousin}.
\begin{align}
    \label{eq:vs_cousin}
    VS_q = \exp\left(\frac{1}{1-q} \log\sum_{i=1}^n \lambda_i^q\right)
\end{align}
Here, $q$ allows users to control the sensitivity to rare (or common) elements, where $q<1$ corresponds to high sensitivity to rare elements. 
The special case of $q=\inf$ forces \ac{VS} to capture the most dominant elements, making it highly sensitive to redundant elements.

\section{Chamfer Distance}
\label{sec:chamfer_dis}
Chamfer Distance (CD) is a geometric metric commonly used to compute the dissimilarity between two sets of points with embeddings. 
Given two sets of sentence embeddings $\cA = \{a_1, a_2, \dots, a_m\}$ and $\cB = \{b_1, b_2, \dots, b_n\}$, CD is defined in \eqref{eq:chamfer}.
\begin{align}
    \label{eq:chamfer}
    \begin{split}
        \text{CD}(A, B) = & \frac{1}{|A|} \sum_{a \in A} \min_{b \in B} \lVert a - b \rVert_2^2 \\
        & + \frac{1}{|B|} \sum_{b \in B} \min_{a \in A} \lVert  b - a \rVert_2^2,
    \end{split}
\end{align}
This metric captures how well each sentence embedding in one set is approximated by the closest embedding in the other set.

\section{Generating Candidate Sets}
\label{sec:app:generator-LLMs}

In this section, we describe further details regarding the high and low quality candidate set generation process.
We use three generator LLMs for this purpose:  \texttt{GPT-4-turbo}~\citep{gpt4}, \texttt{Llama3.1-8b}~\citep{llama3.1}, and \texttt{Qwen 2.5-14b}~\citep{qwen2.5}. 

\begin{figure}[t]
  \centering
  % \rule{7cm}{6cm} % This creates a black rectangle of 6cm x 4cm
  \includegraphics[width=1.0\linewidth]{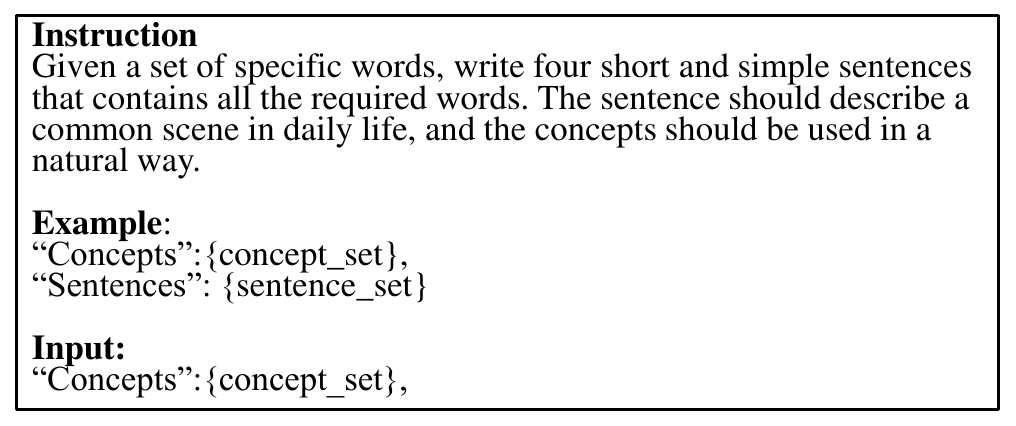}
  \caption{The prompt used to instruct generator LLMs to produce the \textbf{Default} set of sentences.}
  \label{fig:generator-LLM-default-prompt}
\end{figure}

\begin{figure}[t]
  \centering
  %\rule{7cm}{6cm} % This creates a black rectangle of 6cm x 4cm
  \includegraphics[width=1.0\linewidth]{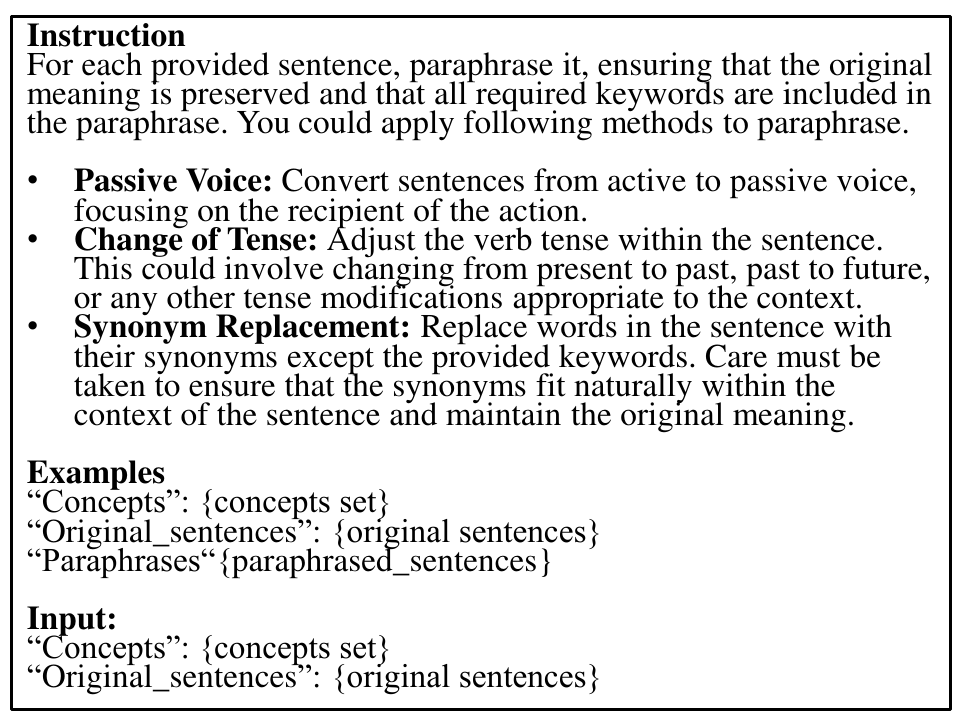}
  \caption{The prompt used to instruct generator LLMs to produce the \textbf{Paraphrased} set of sentences.}
  \label{fig:generator-LLM-para-prompt}
\end{figure}

\begin{figure}[t!]
  \centering
  %\rule{7cm}{6cm} % This creates a black rectangle of 6cm x 4cm
  \includegraphics[width=1.0\linewidth]{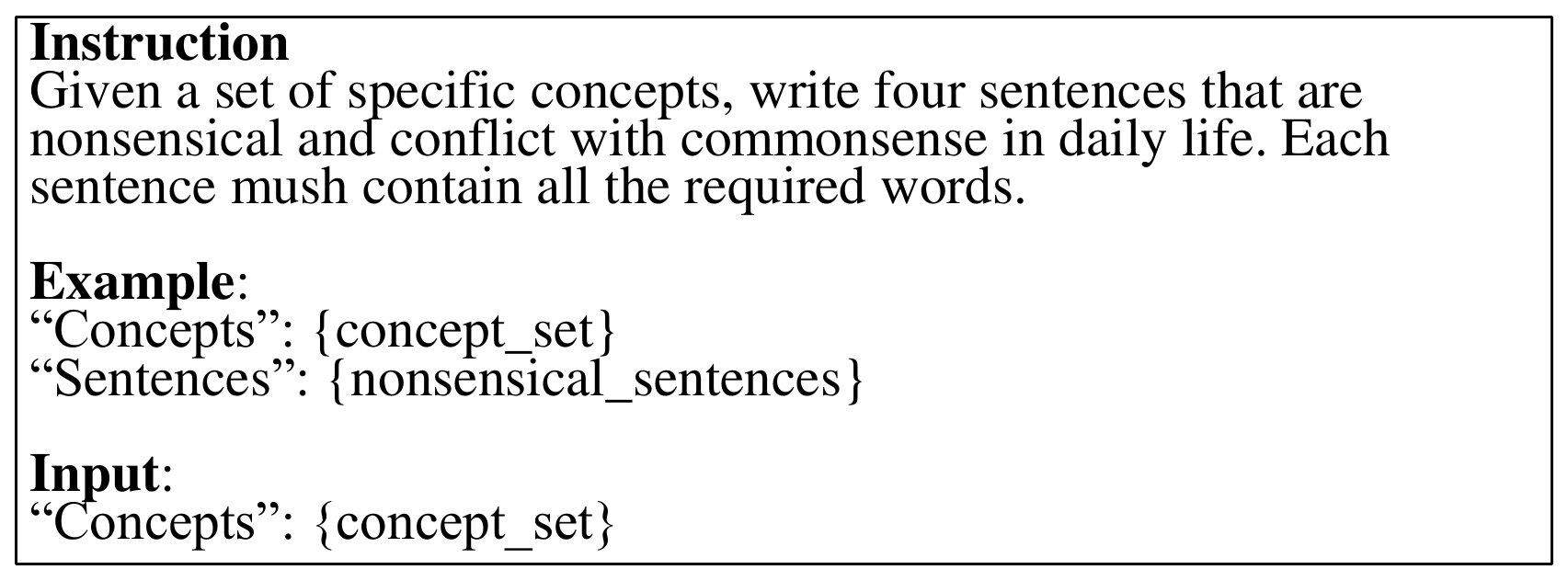}
  \caption{The prompt used to instruct generator LLMs to produce the \textbf{Nonsensical} set of sentences.}
  \label{fig:generator-LLM-nonsensical-prompt}
\end{figure}

To generate the \textbf{Default} set of sentences for each set of input concepts in the CommonGen test cases, we instruct each generator LLM separately with the prompt shown in \autoref{fig:generator-LLM-default-prompt}.
To generate the paraphrase of a given sentence for the \textbf{Para-1}, \textbf{Para-2} and \textbf{Para-3} sets, we instruct the generator LLMs with the prompt shown in \autoref{fig:generator-LLM-para-prompt}.
The instruction to generate nonsensical sentences is shown in \autoref{fig:generator-LLM-nonsensical-prompt}.

An example of the candidate sets of high and low quality sentences generated by the different high and low quality operations are shown in \autoref{tab:samples_candidate_sets}.

\begin{table*}[t]
\centering
\resizebox{\textwidth}{!}{
\begin{tabular}{@{}llll@{}}
\toprule
\textbf{Input concepts} & \{dog, throw, frisbee, catch\} \\
\midrule
\midrule
\multicolumn{2}{@{}c}{\textbf{High-quality candidate sets}} \\
\midrule
\textbf{Default} & "The dog catches the frisbee after I throw it."\\ &"I throw the frisbee and the dog tries to catch it."\\ & "Every time I throw the frisbee, the dog catches it eagerly."\\ &"Watching the dog catch the frisbee after I throw it brings joy." \\
\midrule
\textbf{Para-1} & "After the frisbee is thrown by me, it is caught by the dog." \\ & "The dog catches the frisbee after I throw it."\\
& "I throw the frisbee and the dog tries to catch it." \\ & "Every time I throw the frisbee, the dog catches it eagerly." \\
\midrule
\textbf{Para-2} & "I throw the frisbee and the dog tries to catch it." \\ & "The frisbee is thrown by me, and the dog attempts to catch it."\\
& "Every time I throw the frisbee, the dog catches it eagerly."\\ & "The frisbee is eagerly caught by the dog each time it is thrown by me." \\
\midrule
\textbf{Para-3} & "Every time I throw the frisbee, the dog catches it eagerly."\\ &  "The frisbee is eagerly caught by the dog each time it is thrown by me."\\
& "Eagerly, the frisbee is caught by the dog every time it is thrown by me."\\ & "Watching the dog catch the frisbee after I throw it brings joy." \\
\midrule
\midrule
\multicolumn{2}{@{}c}{\textbf{Low-quality candidate sets}} \\
\midrule
\textbf{Nonsensical} & "The frisbee decided to throw the dog sky-high so it could catch the stars."\\ & "At midnight, every dog gathers to catch the moon while humans throw frisbees into dreams."\\
& "The dog threw a party where frisbees were invited to catch sounds and dances."\\ & "During breakfast, the dog learned to throw sunlight so that the frisbee could catch and store it." \\
\midrule
\textbf{NounShuff} & "The frisbee catches the it after I throw dog."\\ & "dog throw the it and the frisbee tries to catch I."\\
& "Every I frisbee throw the time, the it catches dog eagerly."\\ & "Watching the dog catch the I after frisbee throw it brings joy." \\
\midrule
\textbf{RndShuff} & "catches I dog frisbee it throw the the after."\\ & "I to catch throw dog tries and the it frisbee the."\\
& "the dog I the it frisbee, throw eagerly every time catches."\\ & "the frisbee the catch throw joy I after Watching brings dog it." \\

\bottomrule
\end{tabular}
}
\caption{An example of candidate sets generated by the different high and low quality operations for an input concept set selected from the CommonGen test dataset.}
\label{tab:samples_candidate_sets}
\end{table*}

% Show the quality table of commongen candidate sets.
\subsection{Quality Evaluation}
\label{sec:commongen_quality}
Before evaluating the diversity of each candidate set, we first assess their quality using the official metrics proposed by~\citet{CommonGen} based on \texttt{GPT-4o} for evaluating \ac{LLM} generations. 
Based on these metrics, we classify the candidate sets into high-quality and low-quality groups using the overall quality score.
The quality metrics are defined as follows: 
\begin{description}
    \item[Length:] the number of words on average in the generated sentences.
    \item[Coverage:] the percentage of examples where all given concepts are covered by \ac{LLM} outputs.
    \item[Win\_Tie:] the percentage of examples where GPT-4o prefers the model outputs over the human-written references (or there is a tie).
    \item[Overall Score:] the product of scores on Coverage, and Win\_Tie Rate.
\end{description}
From \autoref{tab:gpt_quality_metrics}, \autoref{tab:qwen2.5_quality_metrics} and ~\autoref{tab:llama3_quality_metrics}, we see that the \textbf{Default} generation achieves the best quality among the candidate sets and the outputs generated by \texttt{GPT-4-turbo} has the best quality among the three models. 
Therefore, we use \texttt{GPT-4-turbo} to show the result in the main paper. 
\texttt{GPT-4-turbo} also has higher win\_tie rate compared with human preference.  
However, as the number of paraphrases increases (e.g. in Para-2 and Para-3), the Win\_Tie decreases. 
This suggests that the CommonGen evaluator implicitly considers diversity as part of its quality evaluation, even though diversity is not explicitly mentioned in the evaluation instructions.
Additionally, the coverage rate declines as the number of paraphrases increases. 
This highlights that generating diverse outputs while maintaining high coverage remains a challenge for \ac{LLM}s, even for state-of-the-art models like \texttt{GPT-4-turbo}.

\section{LLM-based Diversity Template}
\label{sec:app:evaluator-LLM}
\begin{figure*}[t]
\centering
\includegraphics[width=1.0\linewidth]{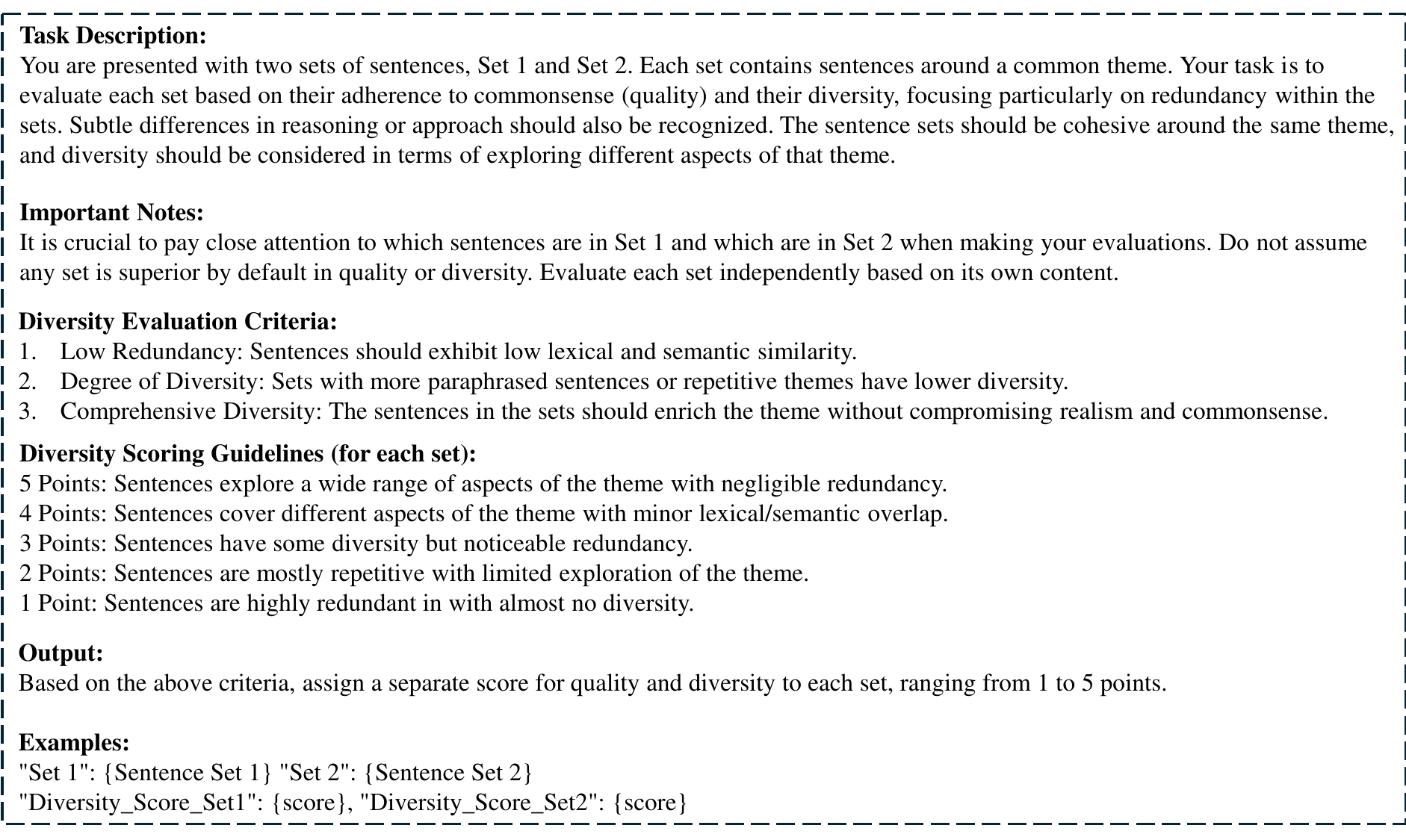}
\caption{Prompt provided to \texttt{GPT-4o} for scoring and comparing two sentence sets. The instructions specify a five-point diversity scale ranging from \emph{highly redundant} (score = 1) to \emph{explore a wide range of aspects of the theme} (score = 5). We also emphasise commonsense consistency and thematic relevance in the instruction. The prompt concludes with a request for a concise output format containing the final scores.}
\label{fig:llm_diversity_template}
\end{figure*}

We use \texttt{GPT-4o} to evaluate and compare the diversity of two sentence sets, using the prompt shown in \autoref{fig:llm_diversity_template}.  
The prompt first defines the diversity criteria to be assessed and explicitly instructs the model to ignore the order of sentences within each set.  
Next, it presents a five-point scoring rubric, where higher scores correspond to greater diversity.  
Finally, the expected output format is specified at the end of the prompt.
 
\section{Temperature Tuning}
\label{sec:temperature_tuning}
To obtain stable diversity ratings, we executed the \ac{LLM} annotator five times per temperature setting and averaged the resulting accuracies. 

As is shown in~\autoref{tab:temp-accuracy},  agreement with human preferences remains high (\,$\ge 77\%$) across all temperatures, confirming the robustness of our annotation pipeline.
The best performance occurs at $Temp\!=\!1.0$, where the average pairwise accuracy peaks at 80.6\%.

\begin{table}[h]
  \centering
  \resizebox{\columnwidth}{!}{
  \begin{tabular}{l|cccccc}
    \toprule
    Temp.          & 0   & 0.2  & 0.4  & 0.6  & 0.8  & 1.0    \\
    \midrule
    Avg.~accuracy & 77.4 & 79.1 & 78.9 & 78.9 & 79.1 & \textbf{80.6} \\
    \bottomrule
  \end{tabular}}
  \caption{Mean pairwise accuracy between the \ac{LLM} annotator and human judgments at different temperatures.}
  \label{tab:temp-accuracy}
\end{table}

\begin{table}[t]
\small
\centering
%\resizebox{\columnwidth}{!}{
\begin{tabular}{p{11mm}cccc}
\toprule
\textbf{Method}           & \textbf{Length} & \textbf{Coverage} & \textbf{Win\_Tie} & \textbf{Overall } \\
\midrule
Default                      & 12.9                      & 86.5                         & 58.7                        & 50.8                        \\
Para-1                  & 12.9                      & 83.1                         & 56.5                        & 47.0                        \\
Para-2                  & 13.8                      & 75.6                         & 43.6                        & 32.9                        \\
Para-3                  & 14.4                      & 75.1                         & 38.5                        & 28.9                        \\
\midrule
Nonsensical               & 15.1                      & 95.4                         & 1.3                         & 1.3                         \\
NounShuff                   & 12.9                      & 85.2                         & 4.9                         & 4.2                         \\
RndShuff                  & 12.9                      & 79.6                         & 0.3                         & 0.2                         \\
\bottomrule
\end{tabular}
%}
\caption{Comparison of length, coverage, win-tie percentage, and overall performance across different methods for the GPT-4-turbo's candidate sets generation. }
\label{tab:gpt_quality_metrics}
\end{table}

% Qwen 2.5-14b model quality
\begin{table}[t]
\centering
\small
%\resizebox{\columnwidth}{!}{
\begin{tabular}{p{11mm}cccc}
\toprule
\textbf{Method}           & \textbf{Length} & \textbf{Coverage} & \textbf{Win\_Tie} & \textbf{Overall} \\
\midrule
Default                      & 13.8            & 62.0              & 44.0              & 27.3             \\
Para-1                   & 14.3            & 55.9              & 38.5              & 21.5             \\
Para-2                   & 15.0            & 46.2              & 30.4              & 14.0             \\
Para-3                 & 15.6            & 46.9              & 27.1              & 12.7             \\
\midrule
Nonsensical               & 18.4            & 73.8              & 1.4               & 1.0              \\
NounShuff                  & 13.8            & 60.9              & 3.7               & 2.2              \\
RndShuff                  & 13.8                      & 55.7                         & 0.1                         & 0.1                         \\
\bottomrule
\end{tabular}
%}
\caption{Comparison of length, coverage, win\_tie percentage, and overall performance across different methods for Qwen2.5-14B’s candidate sets generation.}
\label{tab:qwen2.5_quality_metrics}
\end{table}

% Llama3 quality metrics
\begin{table}[t]
\small
\centering
%\resizebox{\columnwidth}{!}{
\begin{tabular}{p{11mm}cccc}
\toprule
\textbf{Method}           & \textbf{Length} & \textbf{Coverage} & \textbf{Win\_Tie} & \textbf{Overall} \\
\midrule
Default                      & 15.3            & 60.7              & 30.1              & 18.3             \\
Para-1                  & 15.4            & 57.8              & 26.4              & 15.2             \\
Para-2                   & 15.9            & 57.8              & 23.9              & 13.8             \\
Para-3                   & 17.5            & 55.7              & 21.4              & 11.9             \\
\midrule
Nonsensical               & 17.1            & 78.6              & 2.7               & 18.1             \\
NounShuff                  & 15.3            & 59.5              & 2.9               & 1.8              \\
RndShuff                  & 15.3            & 55.6             & 0.1               & 0.1              \\
\bottomrule
\end{tabular}
%}
\caption{Comparison of length, coverage, win\_tie percentage, and overall performance across different methods for the Llama3.1-8B model’s candidate sets generation.}
\label{tab:llama3_quality_metrics}
\end{table}

\section{LLM Reasons about the Diversity Score}
In~\autoref{fig:llm_reason_example}, we show how GPT-4o assigns a diversity score to two different sets of sentences generated from the same input concepts.. A natural question is: \emph{Can an LLM also provide a cogent rationale for its diversity judgments?} 
To explore this, we extend our prompt by appending ``Please include a brief explanation (around 100 words) for your score'' at the end.

\begin{figure*}[t]
\centering
\includegraphics[width=1.0\linewidth]{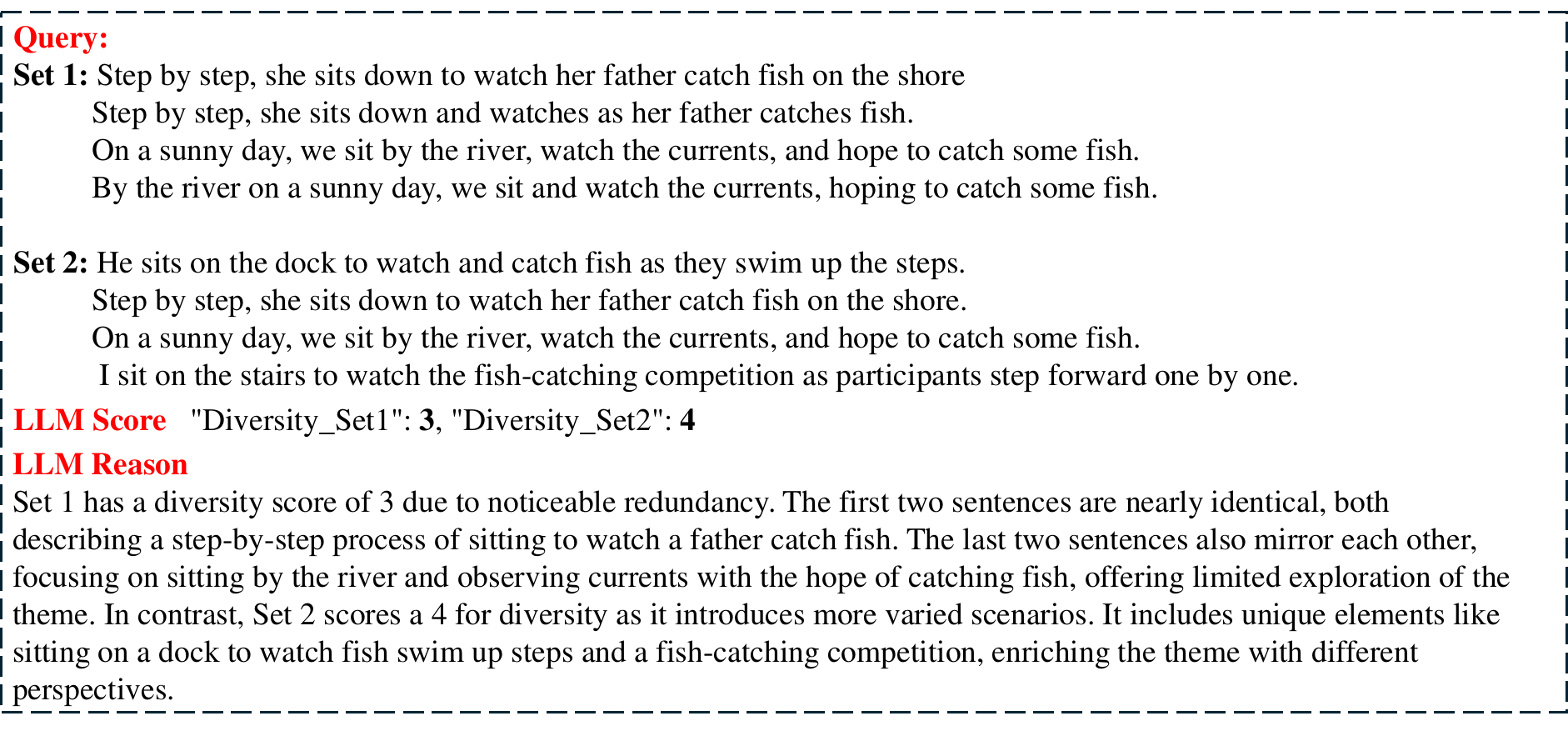}
\caption{GPT-4o’s diversity annotations and accompanying explanation for two sentence sets generation for the source in the CommonGen dataset.}
\label{fig:llm_reason_commongen}
\end{figure*}

\begin{figure*}[t]
\centering
\includegraphics[width=1.0\linewidth]{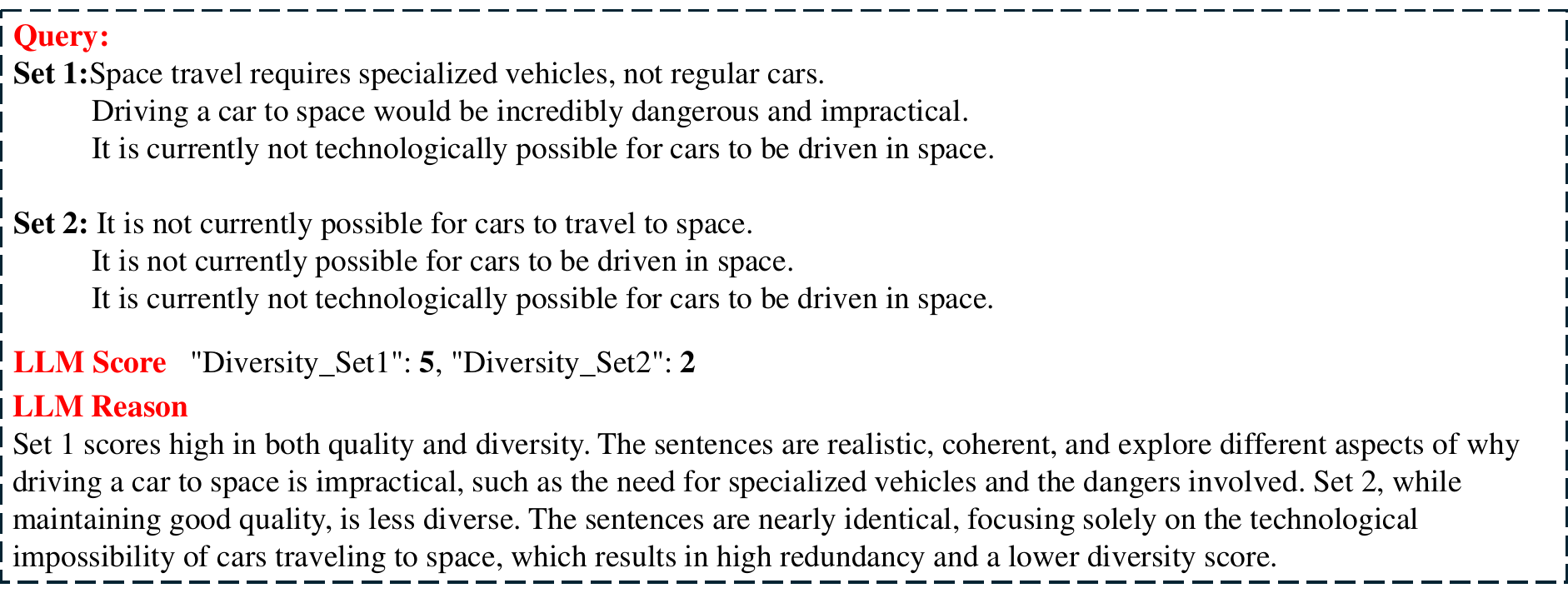}
\caption{GPT-4o’s diversity annotations and accompanying explanation for two sentence sets in the ComVE dataset.}
\label{fig:llm_reason_comve}
\end{figure*}

\begin{figure*}[t]
\centering
\includegraphics[width=1.0\linewidth]{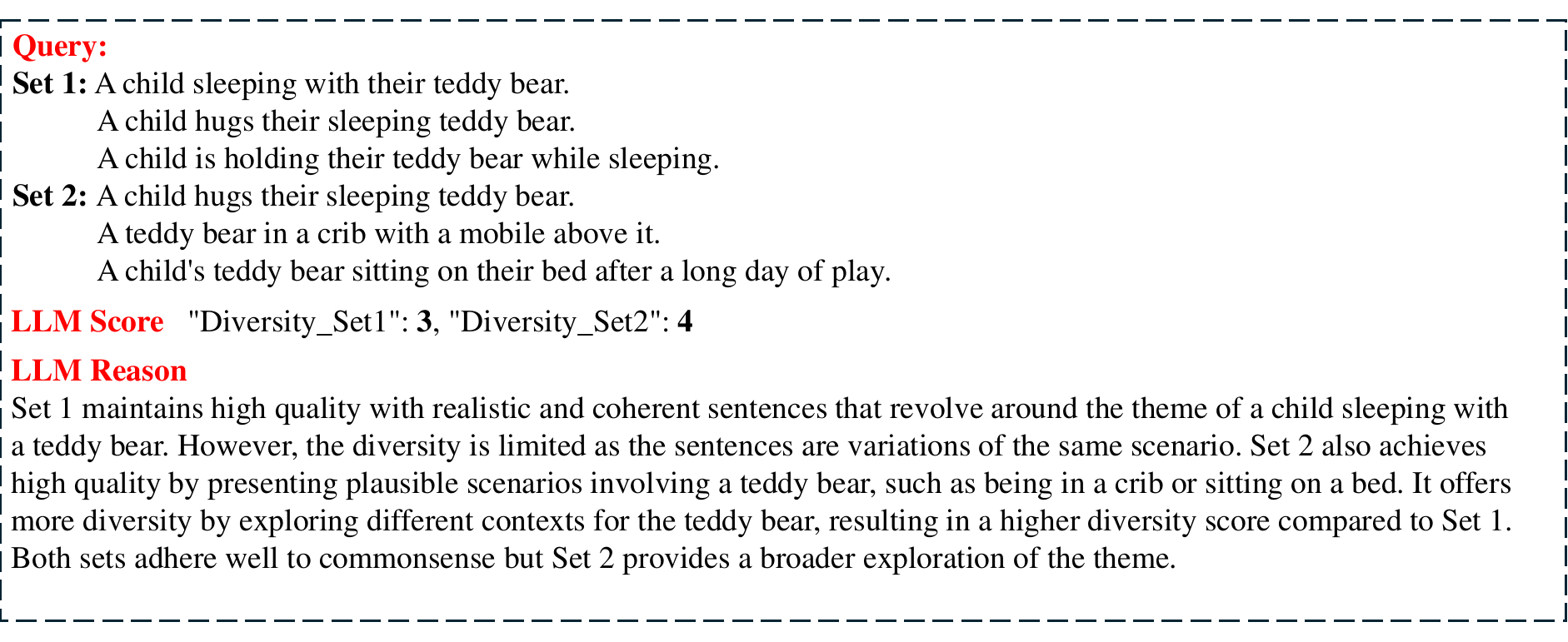}
\caption{GPT-4o’s diversity annotations and accompanying explanation for two sentence sets in the DimonGen dataset.}
\label{fig:llm_reason_dimmongen}
\end{figure*}

Below, we show three representative examples from the CommonGen (\autoref{fig:llm_reason_commongen}), ComVE (\autoref{fig:llm_reason_comve}), and DimonGen (\autoref{fig:llm_reason_dimmongen}) generated sentence set pairs. From the figures, we show that our LLM diversity works on these datasets.

\section{Confidence Intervals}
\label{sec:app:CI}

\begin{figure}[!t]
\centering
\includegraphics[width=1.0\linewidth]{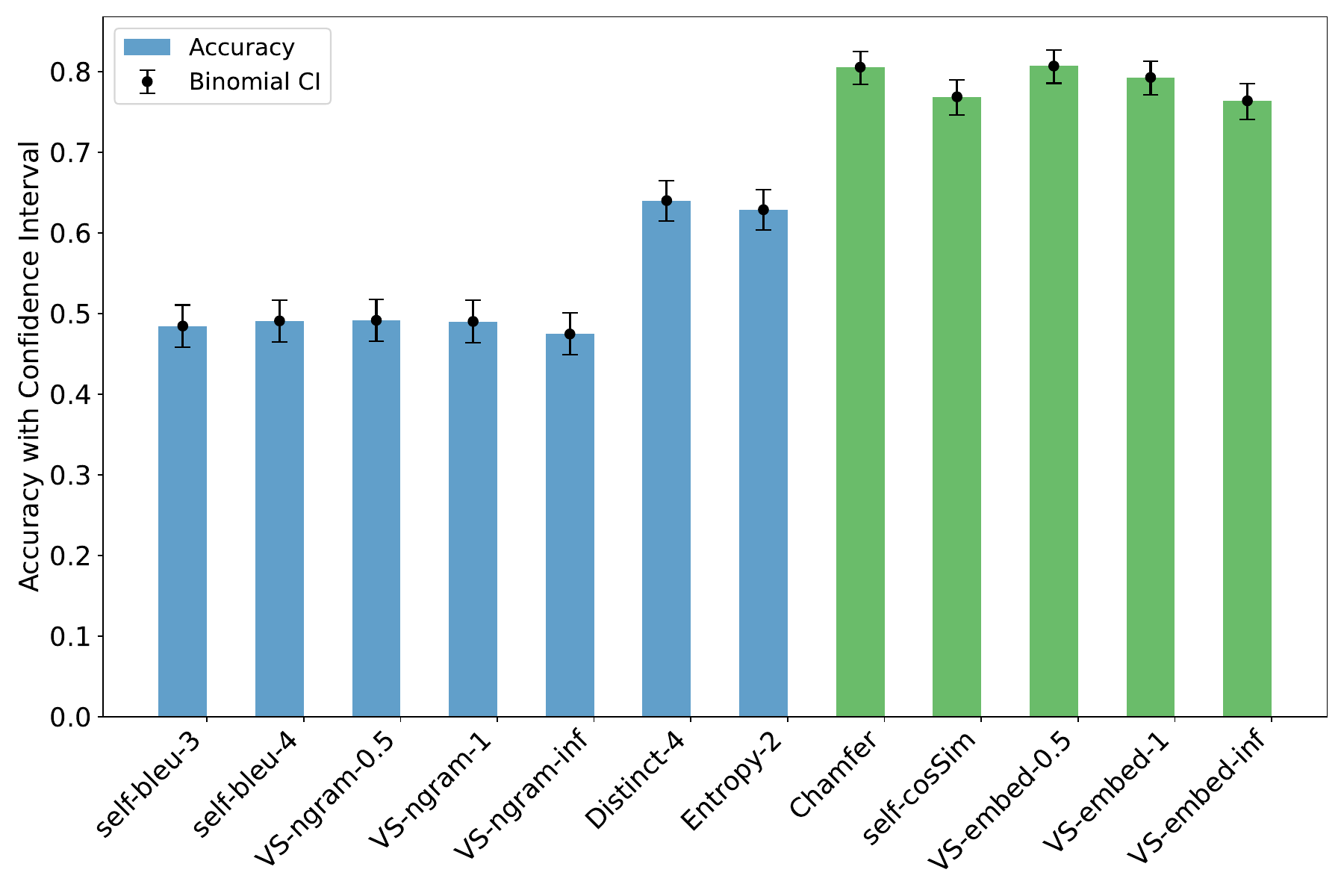}
\caption{Binomial confidence intervals are super-imposed for the accuracies reported by the diversity metrics on the all candidate sentence sets on the CommonGen test dataset}
\label{fig:ci_all}
\end{figure}

\begin{figure}[!t]
\centering
\includegraphics[width=1.0\linewidth]{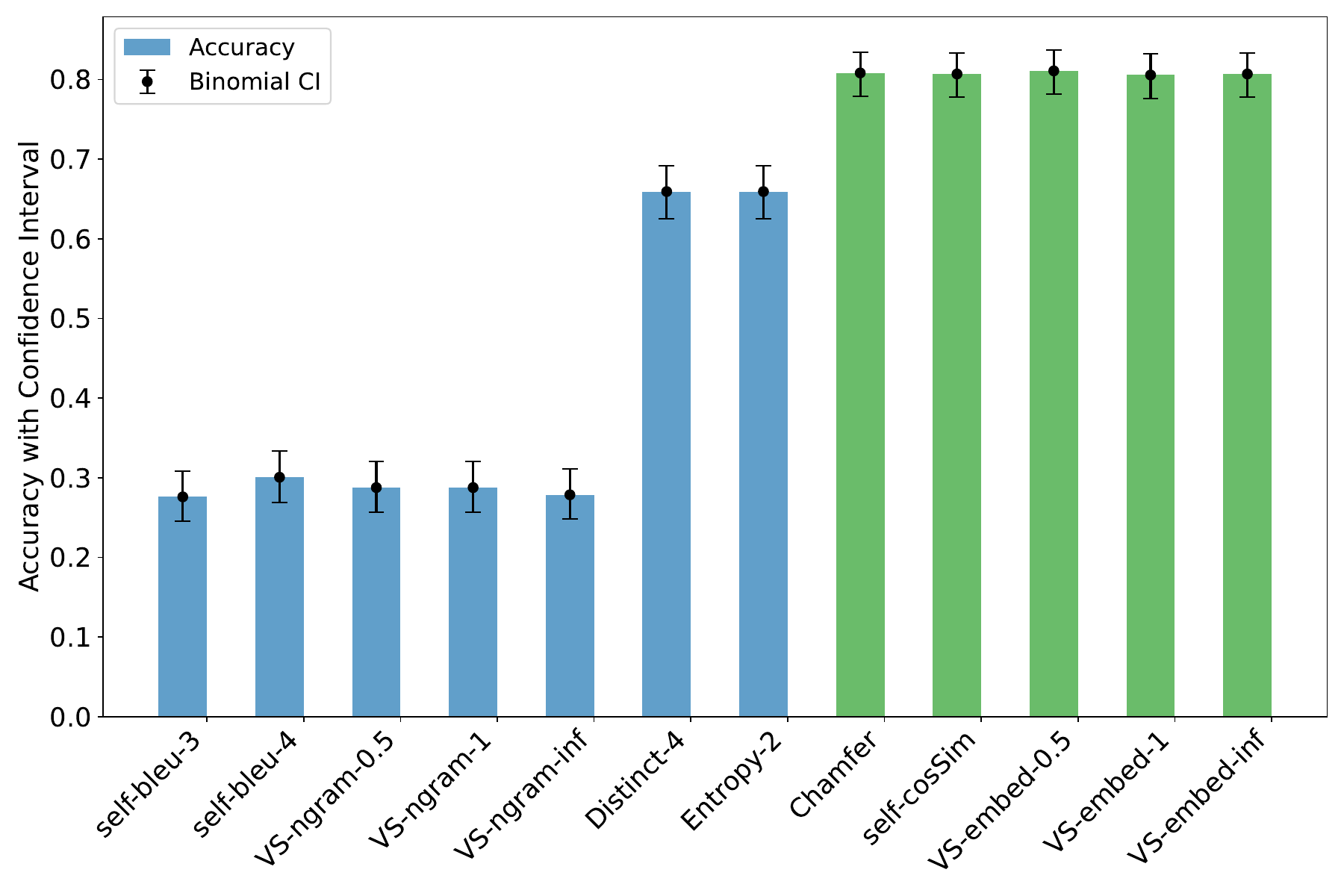}
\caption{Binomial confidence intervals are super-imposed for the accuracies reported by the diversity metrics on the low generation quality candidate sentence sets on the CommonGen test dataset.}
\label{fig:ci_low}
\end{figure}

\begin{figure}[!t]
\centering
\includegraphics[width=1.0\linewidth]{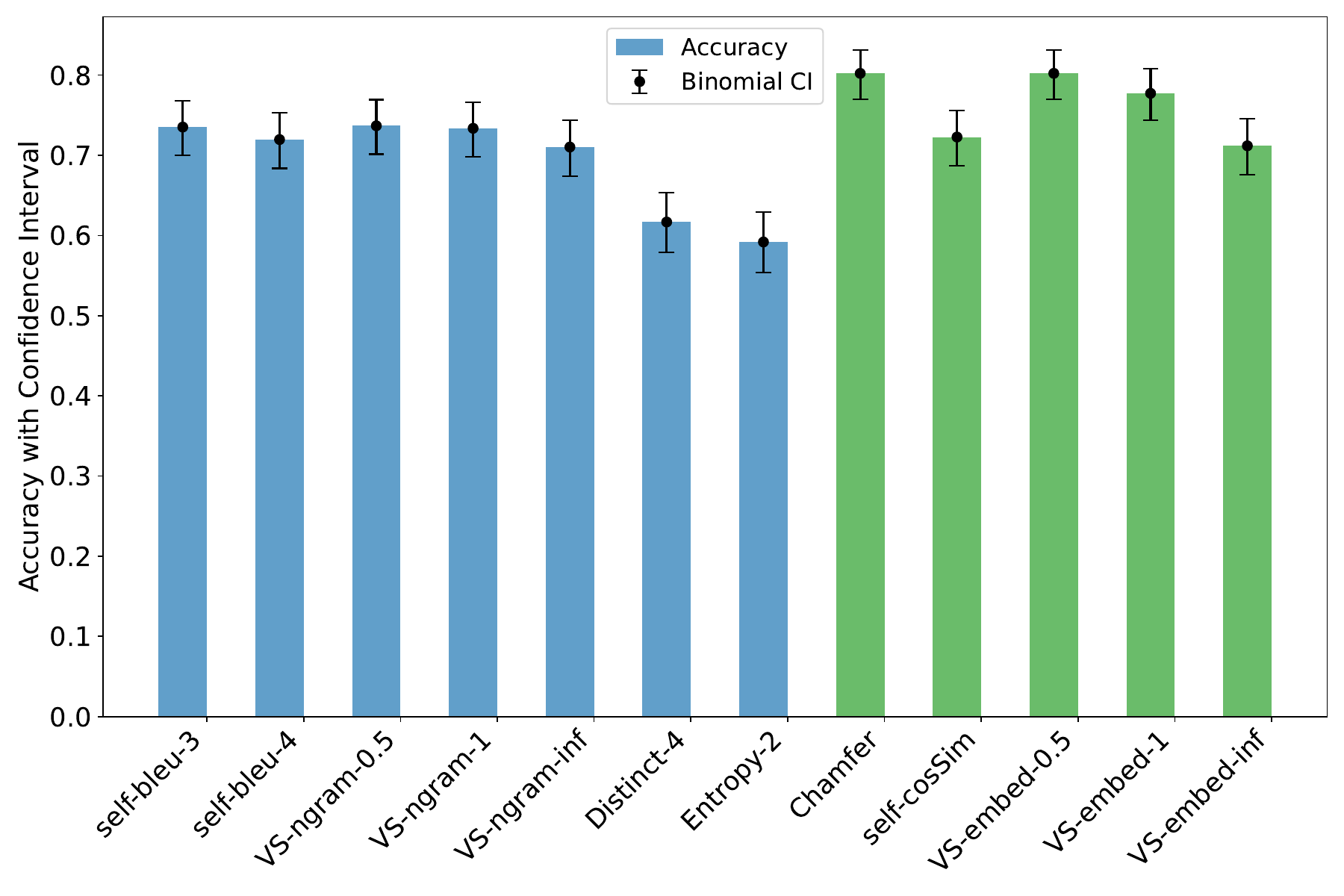}
\caption{Binomial confidence intervals are super-imposed for the accuracies reported by the diversity metrics on the high generation quality candidate sentence sets on the CommonGen test dataset}
\label{fig:ci_high}
\end{figure}

To measure statistical significance for the accuracy scores reported by the different diversity evaluation metrics on the CommonGen dataset, we compute the 95\% binomial confidence intervals using the Clopper-Pearson test~\cite{Clopper:1934} as shown in \autoref{fig:ci_all} for all test cases. 
Additionally, \autoref{fig:ci_high} and \autoref{fig:ci_low} present confidence intervals for the high-quality and low-quality candidate subsets, respectively. 
The bars in blue represent form-based metrics, while the green bars correspond to content-based metrics. 
Across all figures, content-based metrics such as \ac{VS}-Embed-0.5 and Chamfer consistently exhibit higher accuracies with narrower confidence intervals, highlighting their robustness. 
In contrast, form-based metrics such as self-BLEU show lower accuracies and wider intervals, especially in low-quality scenarios. 
These results emphasise the reliability of content-based metrics for evaluating meaningful diversity in \ac{GCR} tasks. 

\section{Further Experiments on ComVE}
\label{sec:further_comve}

To explore the performance of diversity metrics for low quality sentences, we generated low-quality sentence sets on the ComVE dataset, including Nonsensical, NounShuff and RndShuff sentence sets based on the highest-quality generated set by \texttt{Qwen2.5-14B} generator LLM.
We also use \texttt{GPT-4o} as the annotator LLM, and prompt it to provide pairwise diversity judgements to a given pair of sentence sets, resulting in 1,936 test cases. 
The accuracy of each diversity metric is shown in \autoref{tab:comve_futher}. 
We see a clear performance gap between form-based and content-based metrics in this setting as well.
While content-based metrics achieve the highest accuracy, form-based metrics, such as self-BLEU, consistently underperform. 
This experiment further shows the limitations of form-based diversity metrics in capturing meaningful diversity.

\begin{table}[t]
\small
\centering
\begin{tabular}{clc}
\toprule
 & \textbf{Diversity Metric} & \textbf{Accuracy} \\
\midrule
\multirow{7}{*}{\rotatebox{90}{Form}} 
    & self-BLEU-3           & 21.7  \\
    & self-BLEU-4           & 20.5  \\
    & \ac{VS}-ngram-0.5      & 34.7  \\
    & \ac{VS}-ngram-1        & 34.7  \\
    & \ac{VS}-ngram-inf      & 35.2  \\
    & Distinct-4            & 29.3  \\
    & Entropy-2             & 24.0  \\
\midrule
\multirow{5}{*}{\rotatebox{90}{Content}} 
    & Chamfer               & \textbf{38.8}  \\
    & self-cosine           & 38.4  \\
    & \ac{VS}-Embed-0.5      & 38.5  \\
    & \ac{VS}-Embed-1        & 38.5  \\
    & \ac{VS}-Embed-inf      & 38.4  \\
\bottomrule
\end{tabular}
\caption{Accuracy of diversity metrics using low-quality sentence sets generated from the ComVE dataset. We see that form-based metrics perform worse compared to the content-based metrics.}
\label{tab:comve_futher}
\end{table}

\end{document}

%% file: myacronyms.tex
\usepackage{acronym}

\usepackage{etoolbox}
\makeatletter
\newif\if@in@acrolist
\AtBeginEnvironment{acronym}{\@in@acrolisttrue}
\newrobustcmd{\LU}[2]{\if@in@acrolist#1\else#2\fi}

\newcommand{\ACF}[1]{{\@in@acrolisttrue\acf{#1}}}
\makeatother

\acrodef{MLM}[MLM]{Masked Language Model}
\acrodef{LM}[LM]{Language Model}
\acrodefplural{MLM}[MLMs]{\LU{M}{m}asked \LU{L}{l}anguage \LU{M}{m}odels}
\acrodef{LLM}[LLM]{Large Language Model}
\acrodefplural{LLM}[LLMs]{Large Language Models}
\acrodef{SoTA}[SoTA]{state-of-the-art}
\acrodef{ICL}{\LU{I}{i}n-cotext \LU{L}{l}earning}
\acrodef{SCD}{\LU{S}{s}emantic \LU{C}{c}hange \LU{D}{d}etection}
\acrodef{WiC}{Word-in-Context}
\acrodef{ITML}[ITML]{Information-Theoretic Metric Learning}
\acrodef{SDML}[SDML]{Semantic Distance Metric Learning}
\acrodef{GCR}[GCR]{Generative Commonsense Reasoning}
\acrodef{NLG}[NLG]{Natural Language Generation}
\acrodef{NLP}[NLP]{Natural Language Processing}
\acrodef{KG}[KG]{knowledge graph}
\acrodef{CoT}[CoT]{chain-of-thoughts}
\acrodef{VS}[VS]{Vendi Score}
\acrodef{MoE}[MoE]{Mixture of Experts}
\acrodef{KG}[KG]{knowledge graph}